\documentclass[a4paper,fleqn]{cas-dc}

\usepackage[numbers]{natbib}
\def\tsc#1{\csdef{#1}{\textsc{\lowercase{#1}}\xspace}}
\tsc{WGM}
\tsc{QE}
\tsc{EP}
\tsc{PMS}
\tsc{BEC}
\tsc{DE}
\usepackage{subfigure}
\usepackage{stfloats}
\usepackage{booktabs}
\usepackage{ulem}
\usepackage[ruled,linesnumbered]{algorithm2e}
\usepackage{caption}
\usepackage{color}

\begin{document}
\let\WriteBookmarks\relax
\def\floatpagepagefraction{1}
\def\textpagefraction{.001}
\shorttitle{Multi-granularity for knowledge distillation}
\shortauthors{Baitan Shao et~al.}

\title [mode = title]{Multi-granularity for knowledge distillation}                      
\tnotemark[1]
\tnotetext[1]{This work was supported by the National Natural Science Foundation of China under
Grant 61573168}

\author[1]{Baitan Shao}[orcid=0000-0002-3502-4884]
\ead{shaoeric@foxmail.com in}
\author[1]
{Ying Chen}
\cormark[1]
\cortext[cor1]{corresponding author}
\ead{chenying@jiangnan.edu.cn in}

\address[1]{Key Laboratory of Advanced Process Control for Light Industry Ministry of Education, Jiangnan University, Wuxi 214122, China}

\begin{abstract}
Considering the fact that students have different abilities to understand the knowledge imparted by teachers, a multi-granularity distillation mechanism is proposed for transferring more understandable knowledge for student networks. A multi-granularity self-analyzing module of the teacher network is designed, which enables the student network to learn knowledge from different teaching patterns. Furthermore, a stable excitation scheme is proposed to train the student under robust supervision. The proposed distillation mechanism can be embedded into different distillation frameworks, which are taken as baselines. Experiments show the mechanism improves the accuracy by 0.58\% on average and by 1.08\% in the best over the baselines, which makes its performance superior to the state-of-the-arts. It is also exploited that the student's ability of fine-tuning and robustness to noisy inputs can be improved via the proposed mechanism. The code is available at \href{https://github.com/shaoeric/multi-granularity-distillation}{https://github.com/shaoeric/multi-granularity-distillation}.
\end{abstract}


\maketitle

\section{Introduction}
In the past decade, deep learning has played an increasingly important role in computer vision. Huge neural networks are usually used to improve the recognition performance of models. However, such networks have very strict requirements for compute capability and memory, which restricts its development in practical application, especially in industrial fields. In recent years, researchers have carried out studies on model compression, including model pruning \cite{han2015deep, jia2021wrgpruner, wei2021structured}, model quantization \cite{wu2016quantized,hubara2016binarized}, knowledge distillation \cite{hinton2015KD, belal2021knowledge}, light-weight module design \cite{howard2017mobilenets, li2019iirnet} and low-rank decomposition.
\\
Knowledge distillation (KD) is a technique based on transfer learning, whose key idea is to drive a large teacher network and true hard labels to guide a small student network to learn. The soft probabilites output by the teacher can provide more information for the student than true hard labels. In this way, the performance of the student is supposed to be closer to or even better than that of the teacher. In recent years, KD has been frequently used to achieve model compression and improve network performance. For instance, Kumar et al. \cite{kumar2021collaborative} proposed collaborative knowledge distillation (CKD) to predict human actions with missing information under a multi-view setting.\\
Hinton et al. \cite{hinton2015KD} put forward the concept of knowledge distillation for the first time, expecting the student network to fit the outputs of the teacher network to achieve the purpose of model compression. Park et al. \cite{park2019relational} proposed relational knowledge distillation using distance-wise and angle-wise distillation losses in order to transfer mutual relations of data samples. Tian et al. \cite{tian2019crd} introduced contrastive representation into knowledge distillation, which improved greatly the performance of student networks, even surpassing teacher networks. Zhou et al. \cite{zhou2021wsl} investigated the bias-variance tradeoff brought by distillation with soft labels and proposed weighted soft labels to help the network handle the sample-wise bias-variance tradeoff adaptively. Chen et al. \cite{chen2021cross} proposed to automatically assign proper target layer of the teacher for each student layer with an attention mechanism instead of manual association, which optimized the student network with more appropriate guidance.
\\
\begin{figure}
	\centering
	\subfigure[Abstracted knowledge: dog vs bird]{
		\includegraphics[width=0.4\linewidth]{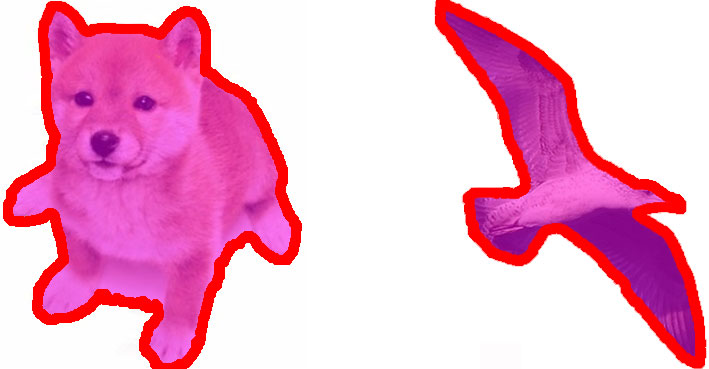}
		\label{fig:AK}
	}
	\hspace{0.1\linewidth}
	\subfigure[Native knowledge: dog vs cat]{
		\includegraphics[width=0.4\linewidth]{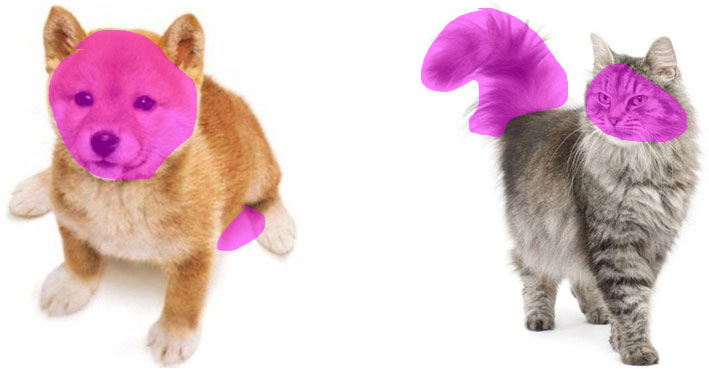}
		\label{fig:NK}
	}
	\subfigure[Detailed knowledge: dog vs dog]{
		\includegraphics[width=0.4\linewidth]{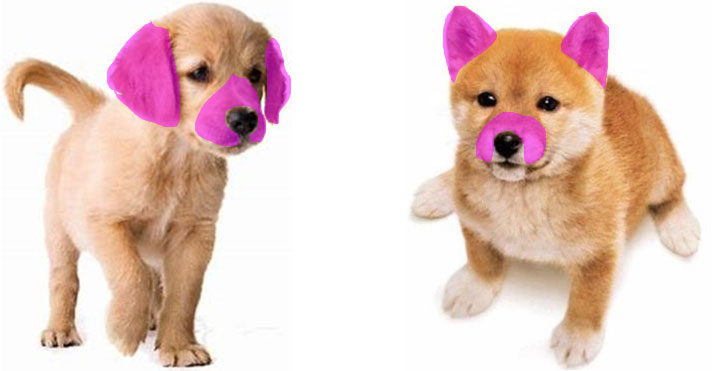}
		\label{fig:DK}
	}%
	\caption{Illustration of multi-granularity knowledge. When distinguishing different categories, different granularities of knowledge and information are needed to pay attention to different aspects of characteristics.}
\label{fig:three_knowledge}
\end{figure}
It is found that the existing methods are to develop an effective but single knowledge for students, while ignoring the fact that students have different ability of knowledge understanding, and it is necessary to teach students in accordance with their aptitude. In the human world, experienced teachers will not only summarize the knowledge to facilitate students' memory of knowledge, but also analyze the knowledge from multiple granularities to help students understand the knowledge better. We denote coarse-grained knowledge as abstracted knowledge, fine-grained knowledge as detailed knowledge and normal-grained knowledge as native knowledge. As illustrated in Figure \ref{fig:three_knowledge}, the three knowledge with different granularity focuses on different discriminative areas for classification. For example, human tend to focus on different object parts when facing differnt classification tasks. A dog and a bird as shown in Figure \ref{fig:AK}, which belong to different phylum, can be distinguished by observing global information such as the outlines of the two objects. A dog and a cat as shown in Figure \ref{fig:NK}, which belong to different family while to the same order, can be distinguished from each other by observing their faces. However, much more detailed information are needed, such as subtle differences in their ears and mouths, when distinguishing different dogs with the same species as shown in Figure \ref{fig:DK}.
\\
Based on the above observation, a multi-granularity knowledge disillation mechanism is proposed for transferring comprehensive knowledge. A multi-ganulairty self-analyzing module designed to construct knowledge of multiple granularities for the teacher and enable the student to learn from different teaching patterns. Furthermore, a stable excitation scheme is proposed to provide robust supervison on the student, which prevents the loss of the student network from violent oscillation during the early optimization period. The proposed distillation mechanism can be embedded into different distillation frameworks, furtherly improving the performance of the state-of-the-arts. Experiments also show that the mechanism can improve the student networks' fine-tuning ability and robustness to noisy inputs.
The contribution of the work can be summarized as follows:
\begin{itemize}
	\item[$\bullet$] A multi-granularity knowledge distillation mechanism is proposed for transferring more understandable knowledge for students. To our best knowledge, it is the first time to introduce the multi-granularity property of knowledge into distillation framework.
	\item[$\bullet$] A multi-granularity self-analyzing module is designed for the teacher network, in which the multi-granularity knowledge is constructed from the native one via fully connected branches of different dimensions.
	\item[$\bullet$] A stable excitation scheme is proposed, which ensembles multi-granularity knowledge for robust supervision on students, avoiding violent oscillation during the early distillation.
	\item[$\bullet$] The mechanism can be embedded into different distillation frameworks, which improves the accuracy by 0.58\% on average and by 1.08\% in the best over the baselines. The performance is superior to the state-of-the-arts.
	\item[$\bullet$] The proposed mechansim can also improve students' ability of fine-tuning and robustness to noisy inputs, which is proved by the experiments.
\end{itemize}

\section{Related work}
In the past few years, many effective methods and theories of knowledge distillation have been proposed, which can be mainly divided into response-based, feature-based and relation-based knowledge distillation \cite{gou2021knowledge}.
\\
Response-based knowledge distillation transfers teacher's knowledge at logits. Buciluˇa et al. \cite{bucilua2006model} introduced the idea of logit matching to transfer the knowledge of large and cumbersome models to smaller and faster models with little performance degradation.
Based on this idea, Hinton et al. \cite{hinton2015KD} introduced the concept of temperature to transfer more knowledge information of a teacher into a student network, avoiding logits degenerating into onehot labels. 
Zhou et al. \cite{zhou2021wsl} investigated the bias-variance tradeoff brought by distillation with soft labels and proposed weighted soft labels to help the network handle the sample-wise bias-variance tradeoff adaptively. 
\\
Feature-based knowledge distillation transfers teacher's knowledge via intermediate representations. Heo et al. \cite{heo2019comprehensive} proposed a feature-based distillation method using network transforms, specific distillation feature position and a partial $L_2$ distance function to preserve useful information as much as possible and skip redundant information. 
Chen et al. \cite{chen2021cross} proposed cross-layer distillation to automatically assign proper target layer of the teacher for each student layer with an attention mechanism instead of manual association, which optimized the student network with more appropriate guidance. 
\\
Relation-based knowledge distillation transfers teacher's knowledge on sample relations.  Park et al. \cite{park2019relational} proposed distance-wise and angle-wise distillation losses, aiming to transfer structured information through the mutual relations of data samples, which provided more ideas for the follow-up research of relation-based knowledge distillation. 
With the significant performance improvement of contrastive learning in unsupervised learning, Tian et al. \cite{tian2019crd} introduced the idea of contrastive learning into knowledge distillation and proposed contrastive representation distillation, which significantly improved the generalization ability of student networks and even surpassed the of teacher networks. In the meantime, Tian et al. \cite{tian2019crd} also demonstrated that contrastive representation distillation had good transfer ability through a series of experiments, including dataset transfer and cross-modal transfer.
\\
In addition, online distillation is also a research focus of knowledge distillation. In the training phase, multiple sub-networks are used to distill each other without pretraining a teacher network.  Zhang et al. \cite{zhang2018deep} presented deep mutual learning (DML), which was simpler and more effective than conventional distillation methods. At the same time, they also discussed why DML worked and why DML was not supposed to be regarded as "the blind lead the blind".
Based on the DML framework,  Guo et al. \cite{guo2020online} proposed a knowledge distillation method via collaborative learning, termed KDCL, which can steadily improve the generalization ability of networks. Unlike DML, KDCL uses the ensemble predictions of all sub-networks as a soft target to supervise the distillation of each sub-network.
Zhang et al. \cite{zhang2021adversarial} proposed adversarial co-distillation networks (ACNs) to generate the divergent exsamples and further facilitate learning, which enhanced the quality of the co-distilled dark knowledge.

\section{Multi-granularity distillation mechanism}
\begin{figure}[]
	\centering
	\includegraphics[width=\linewidth]{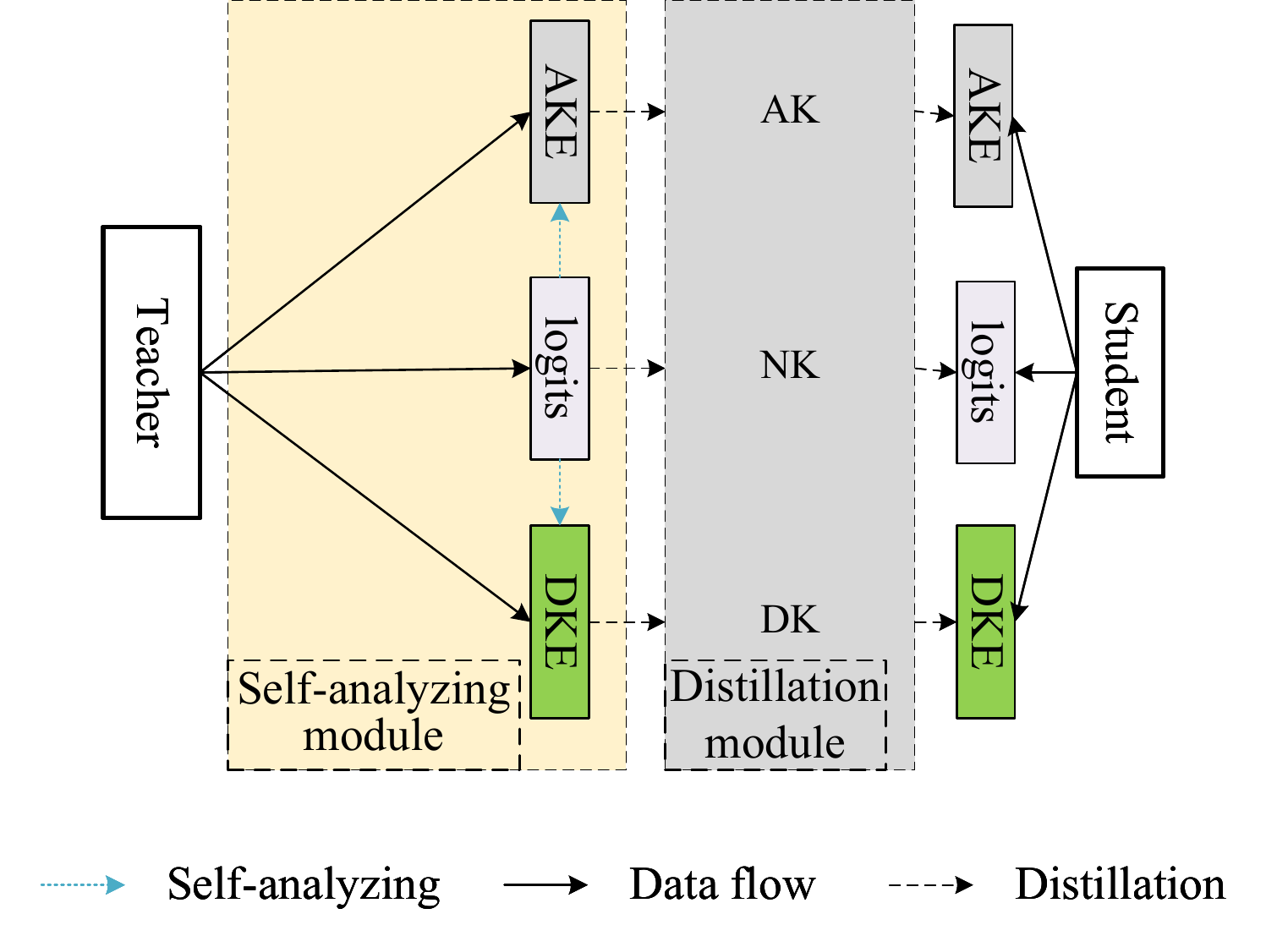}
	\caption{Diagram of multi-granularity distillation mechanism. AK, NK and DK denote abtracted, native and detailed knowledge, respectively. AKE denotes AbstractedKE and DKE denotes DetailedKE.}
	\label{fig:overall}
\end{figure}
Different from previous works, in which a single knowledge is transfered from teachers to students, the multi-granularity distillation mechanism aims to explore multiple granularity of teacher's knowledge and help students learn from different aspects. The diagram of the mechanism is illustrated in Figure \ref{fig:overall}, which includes two modules: multi-granularity self-analyzing module and distillation module. For the multi-granularity self-analyzing, abstracted knowledge encoder (AbstractedKE) and detailed knowledge encoder (DetailedKE) are constructed, which are expected to produce abstracted knowledge and detailed knowledge respectively. In the figures and formulas in this paper, AbstractedKE and DetailedKE are abbreviated to AKE and DKE for legibility. For the distillation, two schemes are designed, namely granularity-wise distillation and stable excitation distillation.\\
Formulaically, given a trained teacher network $T$ and a student network $S$, given training data with $N$ sample pairs $(x_i, y_i)_{i=1}^N$ from $C$ categories, where $x_i$ is the $i$-th input data and $y_i$ is the corresponding ground truth, $F_T(x_i)$ and $F_S(x_i)$ are denoted as output at logits of $T$ and $S$ for $x_i$ respectively. The loss function of a conventional distillation Hinton knowledge distillation (HKD) \cite{hinton2015KD} is as follows:
\begin{equation}
	L_{H}(F_T,F_S,\tau) =\frac{1}{N}\tau ^2 \sum_{i=1}^{N} KL\left(\phi \left( \frac{F_T(x_i)}{\tau} \right) ,\phi \left( \frac{F_S(x_i)}{\tau} \right) \right),    
	\label{L_H}
\end{equation}
where $\tau$ is denoted as the distillation temperature, $KL(\cdot)$ and $\phi (\cdot)$ denote KL divergence \cite{kullback1951information} and softmax function respectively.
\subsection{Multi-granularity self-analyzing module}
\label{Self-analyzing module}
\begin{figure}[]
	\centering
	\includegraphics[width=0.9\linewidth]{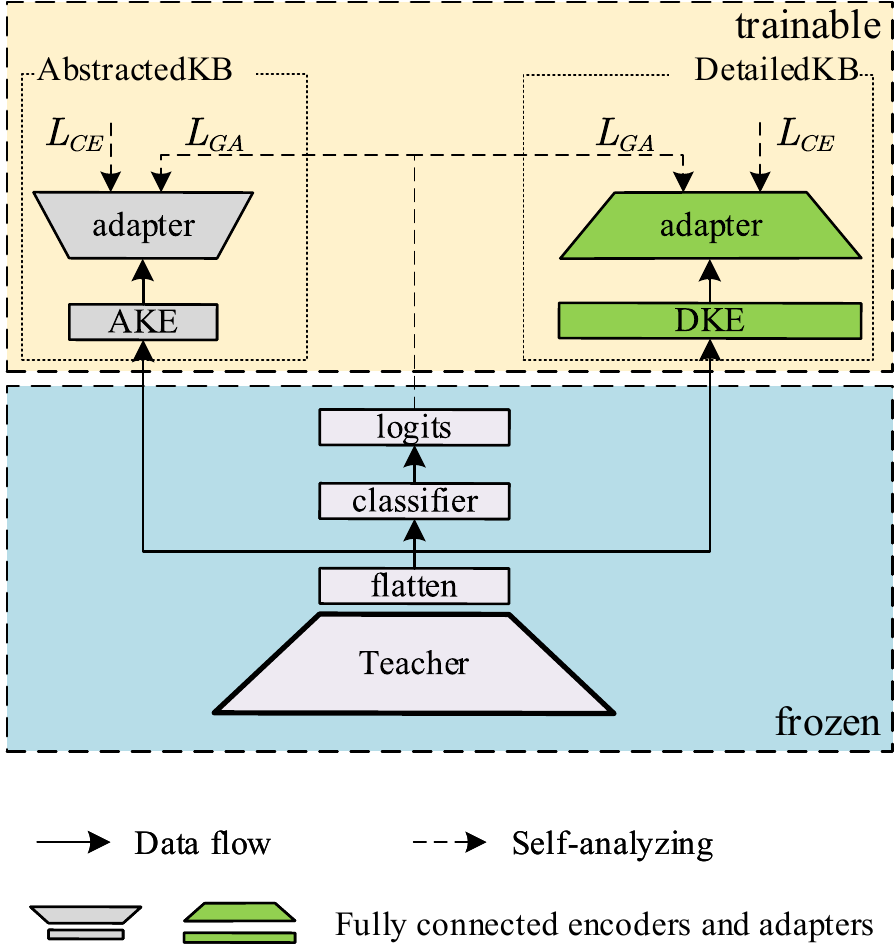}
	\caption{Self-analyzing module. Images are fed into the teacher network and then the flatten representations extracted by backbone are input into the teacher's classifier, AbstractedKB and DetailedKB. When the two branches are trained under the supervision of logits and ground truth, the parameters of the backbone and the classifier should be frozen.}
	\label{fig:self-analyzing}
\end{figure}
\begin{figure*}
	\centering
	\includegraphics[width=0.6\linewidth]{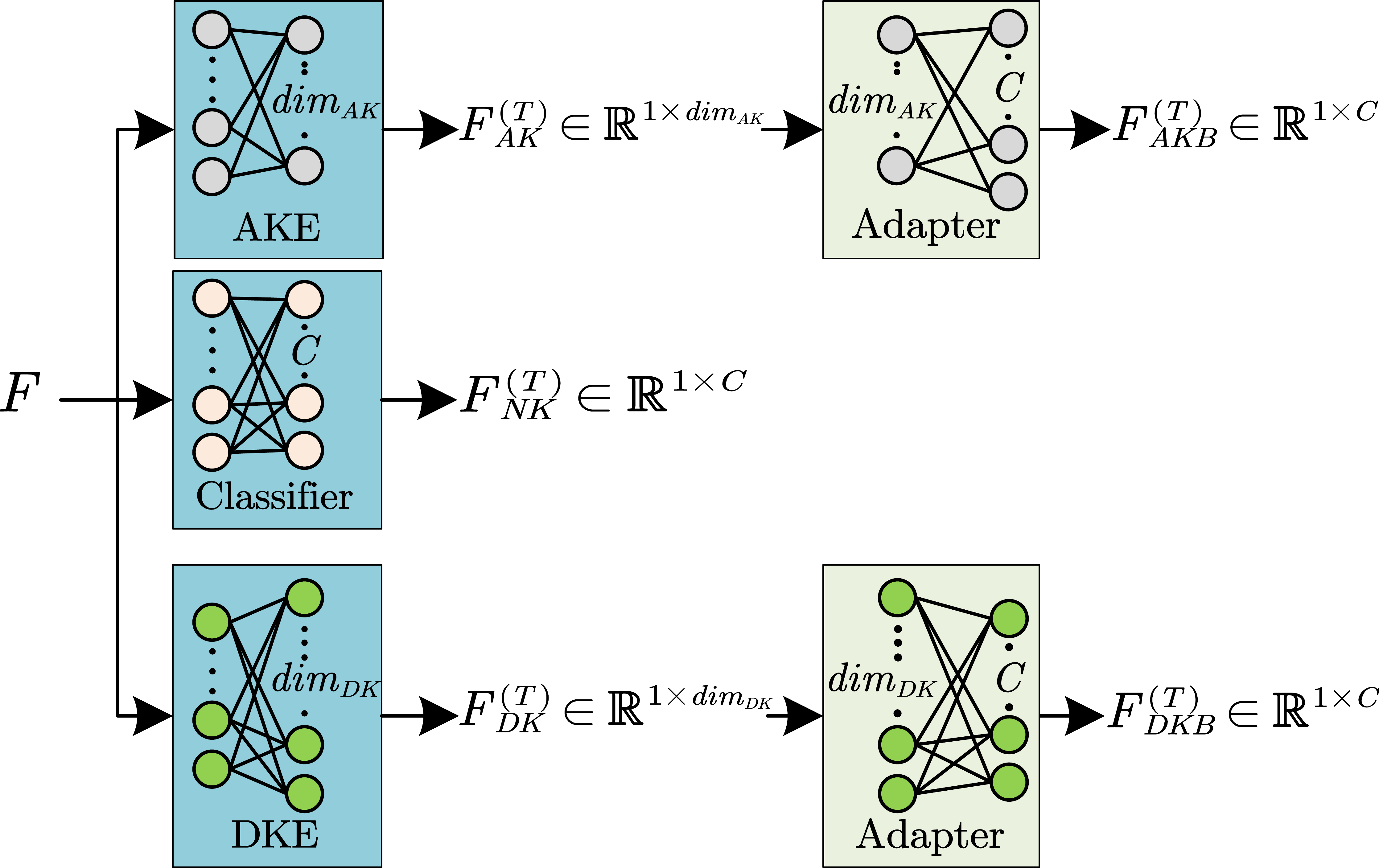}
	\caption{Diagram of the encoders and adapters in granularity self-analyzing module, where mini-batch is set to 1. $F$ denotes the flatten representation of the teacher, $F_{AK}^{(T)}$, $F_{NK}^{(T)}$ and $F_{DK}^{(T)}$ denote three kinds of multi-granularity knowledge outputs, and $F_{AKB}^{(T)}$ and $F_{DKB}^{(T)}$ represent the outputs of the two adapters to align $F_{NK}^{(T)}$ in order to optimize AKE and DKE. }
	\label{fig:encoder_adapter}
\end{figure*}

\begin{figure*}
	\centering
	\subfigure[granularity-wise distilltion]{
		\centering
		\includegraphics[height=0.26\textheight]{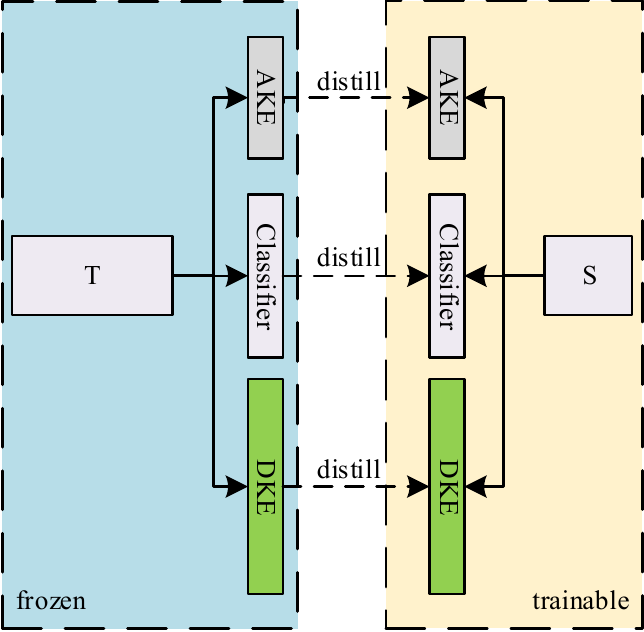}
		\label{fig:granularity-wise distill}
	}%
	\hspace{0.05\linewidth}
	\subfigure[stable excitation distillation]{
		\centering
		\includegraphics[height=0.26\textheight]{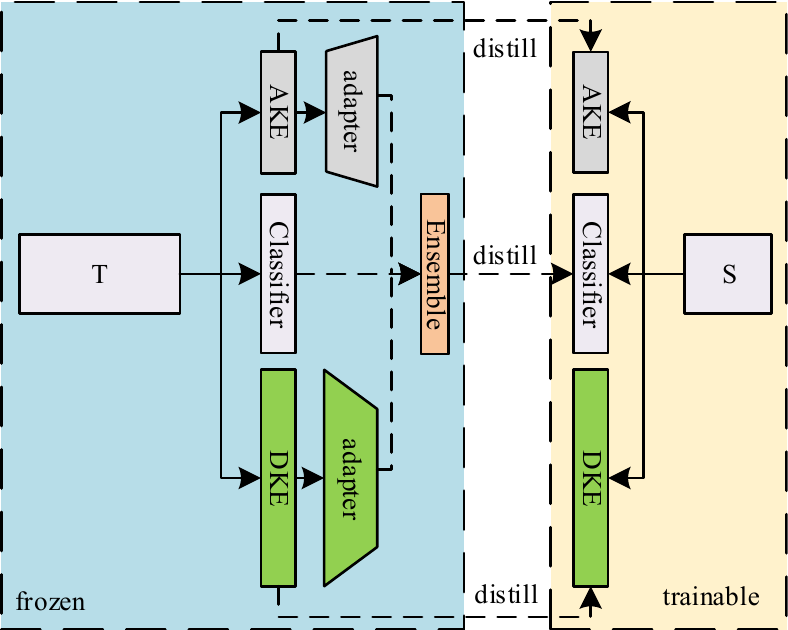}
		\label{fig:stable excitation mechanism}
	}%
	\centering
	\caption{The two subfigures show the structures of granularity-wise distillation and stable excitation distillation respectively. Granularity-wise distillation module distills according to the dimension of the encoders. On the basis of granularity-wise distillation, stable excitation distillation scheme integrates the logits and the outputs of AbstractedKB and DetailedKB to construct stable supervision.}
\end{figure*}
The multi-granularity self-analyzing module aims to construct knowledge with various granularities by optimizing the two encoders under the supervision of the trained teacher's native knowledge at logits in turn, where the backbone and the classifier of the teacher are frozen, and the abstracted knowledge branch (AbstractedKB) and the detailed knowledge branch (DetailedKB) are trainable, as illustrated in Figure \ref{fig:self-analyzing}. In the formulas in this paper, AbstractedKB and DetailedKB are abbreviated to AKB and DKB for legibility.\\
For an input image, i.e. minibatch is 1, it is fed into $T$ to get flatten representation $F$. AbstractedKE and DetailedKE are designed as a fully connected layer structure like the classifier, whose input dimensions are equal to the input dimension of the classifier. As illustrated in Figure \ref{fig:encoder_adapter}, $F$ is fed into AbstractedKE, classifier and DetailedKE of $T$ respectively in order to construct multi-granularity knowledge of $T$: $F_{AK}^{(T)}\in \mathbb{R}^{1\times dim_{AK}}$, $F_{NK}^{(T)}\in \mathbb{R}^{1\times C}$ and $F_{DK}^{(T)}\in \mathbb{R}^{1\times dim_{DK}}$
, where AK, NK and DK denote abstracted knowledge, native knowledge and detailed knowledge respectively, and $dim_{AK}$ and $dim_{DK}$ are denoted as the output dimensions of AbstractedKE and DetailedKE respectively, satisfying the following relationship:
\begin{equation}
	dim_{AK} \textless C \textless dim_{DK}.
\end{equation}
It is expected that the low-dimensional AbstractedKE can analyze and construct an essential coarse-grained knowledge, focusing on the global characteristics of an image, which is similar to supervised dimensionality reduction methods such as LDA \cite{izenman2013linear}, while the high-dimensional DetailedKE can obtain more detailed and fine-grained knowledge, focusing on the detailed features.\\
In order to align the dimensions of AbstractedKB and DetailedKB with the dimension of native knowledge at logits for training, a fully connected adapter is attached to AbstractedKE and DetailedKE respectively, as illustrated in Figure \ref{fig:encoder_adapter}. The input dimensions of the adapter's weight in the AbstractedKB and DetailedKB are $1\times dim_{AK}$ and $1\times dim_{DK}$ respectively and their output dimensions are both $C$, with the $T$'s AbstractedKB output $F_{AKB}^{(T)}\in \mathbb{R}^{1\times C}$ and DetailedKB output $F_{DKB}^{(T)}\in \mathbb{R}^{1\times C}$.\\
The objective function of the module includes two parts, namely granularity analysis loss and cross entropy loss, in which the former is designed for the following requirements: \\
(1) It enables the two branches to capture the teacher's native knowledge as accurately as possible instead of destroying the structure of native knowledge, which can be achieved via HKD \cite{hinton2015KD} to make the output distribution of the two branches close to native knowledge distribution or be treated as a regression problem. \\
(2) It needs temperature parameters to control the abundance of knowledge gained. Lower temperature helps decompose abstracted knowledge and reduce the interference of irrelevant information, while higher temperature produces richer detailed knowledge and strengthens the network's understanding of input. In this work, temperature parameters \\ $\left\{ \tau _{AKB},\tau _{DKB}|\tau _{AKB}<\tau _{DKB} \right\} $ are set for granularity analysis on AbstractedKB and DetailedKB respectively. Therefore the granularity analysis loss function $L_{GA}$ of branch $b\in \{\mathrm{AbstractedKB},\mathrm{DetailedKB}\}$ is denoted as:
\begin{equation}
	L_{GA}(F_{NK}^{(T)}, F_{b}^{(T)}, \tau_b) = L_{H}(F_{NK}^{(T)}, F_{b}^{(T)}, \tau_b),
	\label{L_GA}
\end{equation}
where $F_{b}^{(T)}\in \{ F_{AKB}^{(T)}, F_{DKB}^{(T)}\}$, $\tau_b \in \{ \tau_{AKB}, \tau_{DKB}\}$ denotes the temperature parameter of branch $b$, $L_H$ refers to Equation \eqref{L_H}. \\
In addition, the two branches should also be supervised by cross entropy loss to correct potential errors in native knowledge at logits, which can be denoted as:
\begin{equation}
	L_{CE}\left( F_b^{(T)}, y\right)=\frac{1}{N}\sum_{i=1}^N{-\sum_{c=1}^C{y_{ic}\log \left( \phi \left( F_{b}^{(T)} \right) _{ic} \right)}},
\end{equation}
where $c$ denotes the $c$-th category. The objective function of multi-granularity self-analyzing module of branch $b$ is denoted as:
\begin{equation}
	L_{SA}\left( F_{NK}^{(T)}, F_b^{(T)},\tau _b \right)= L_{GA}\left( F_{NK}^{(T)}, F_b^{(T)},\tau _b \right) + L_{CE}\left( F_b^{(T)}, y\right). 
	\label{L_SA}
\end{equation}
In order to clarify the multi-granularity self-analyzing, the process is given in Algorithm \ref{algorithm:Multi-granularity self-analyzing}.
\begin{algorithm}
	\caption{Multi-granularity self-analyzing}\label{algorithm:Multi-granularity self-analyzing}
	\KwIn{Pretrained and frozen teacher network $T$, abstracted knowledge dimension $dim_{AK}$, detailed knowledge dimension $dim_{DK}$.}
	\KwOut{Optimal self-analyzed teacher network $T_{SA}$ with multi-granularity knowledge}
	Construct fully connected AbstractedKB with AbstractedKE and adapter, whose output dimensions are $dim_{AK}$ and $C$ respectively;\\
	Construct fully connected DetailedKB with DetailedKE and adapter, whose output dimensions are $dim_{DK}$ and $C$ respectively;\\
	\While{not converged}
	{
		Feed the image batch into the backbone of $T$;\\
		Calculate the flatten representation $F$ of $T$;\\
		Feed $F$ into classifier, AbstractedKB and DetailedKB respectively;\\
		Calculate the corresponding output $F_{NK}^{(T)}$, $F_{AKB}^{(T)}$ and $F_{DKB}^{(T)}$;\\
		Optimize the AbstractedKB and DetailedKB under the supervision of $F_{NK}^{(T)}$ and ground truth via Equation \eqref{L_SA} in turn.
	}
\end{algorithm}

\subsection{Distillation module}
After the teacher network completes the multi-granularity self-analyzing, we get a self-analyzed teacher network $T_{SA}$ with multi-granularity knowledge. Afterwards, distillation needs to be conducted, where $T_{SA}$ is frozen, and the native student network and its multi-granularity branches are trainable.\\
In this work, two schemes are designed, namely granularity-wise distillation scheme (GWD) and stable excitation distillation scheme (SE). For granularity-wise distillation, the multi-granularity knowledge is transferred to the student. For stable excitation distillation scheme, an ensemble module is presented for robust teacher supervision transferred to students. \\
For convenience, the proposed distillation mechanism with multi-granularity self-analyzing module and GWD is abbreviated as MAG, and the mechanism with multi-granularity self-analyzing module and SE is abbreviated as MAS.
\paragraph{Granularity-wise distillation}
\label{Granularity-wise distillation}
The AbstractedKE and the DetailedKE of the student are randomly initialized, and the student and its two encoders can be optimized jointly under the supervision of $T_{SA}$, where the AbstractedKE, classifier and DetailedKE of the teacher distill knowledge into the AbstractedKE, classifier and DetailedKE of the student respectively, as illustrated in Figure \ref{fig:granularity-wise distill}.  Similar with the relevant identification of the teacher, the outputs of the student's AbstractedKE, classifier and DetailedKE are denoted as $F_{AK}^{(S)}$, $F_{NK}^{(S)}$ and $F_{DK}^{(S)}$ respectively. The objective function is denoted as:
\begin{equation}
		L_{GWD}=\sum_{k}L_H(F_{k}^{(T_{SA})}, F_{k}^{(S)}, \tau_k) + L_{KD},
	\label{L_{GWD}}
\end{equation}
where $k\in \left\{\mathrm{AK}, \mathrm{DK}, \mathrm{NK}\right\}$, $L_{KD}$ can be the loss of any KD method. With the help of $L_{KD}$, the proposed module can be embedded into SOTA distillation frameworks, which furtherly improves the performance of SOTA distillation methods.
\paragraph{Stable excitation distillation}
\label{Stable excitation distillation}
\begin{figure}[]
	\centering
	\includegraphics[width=\linewidth]{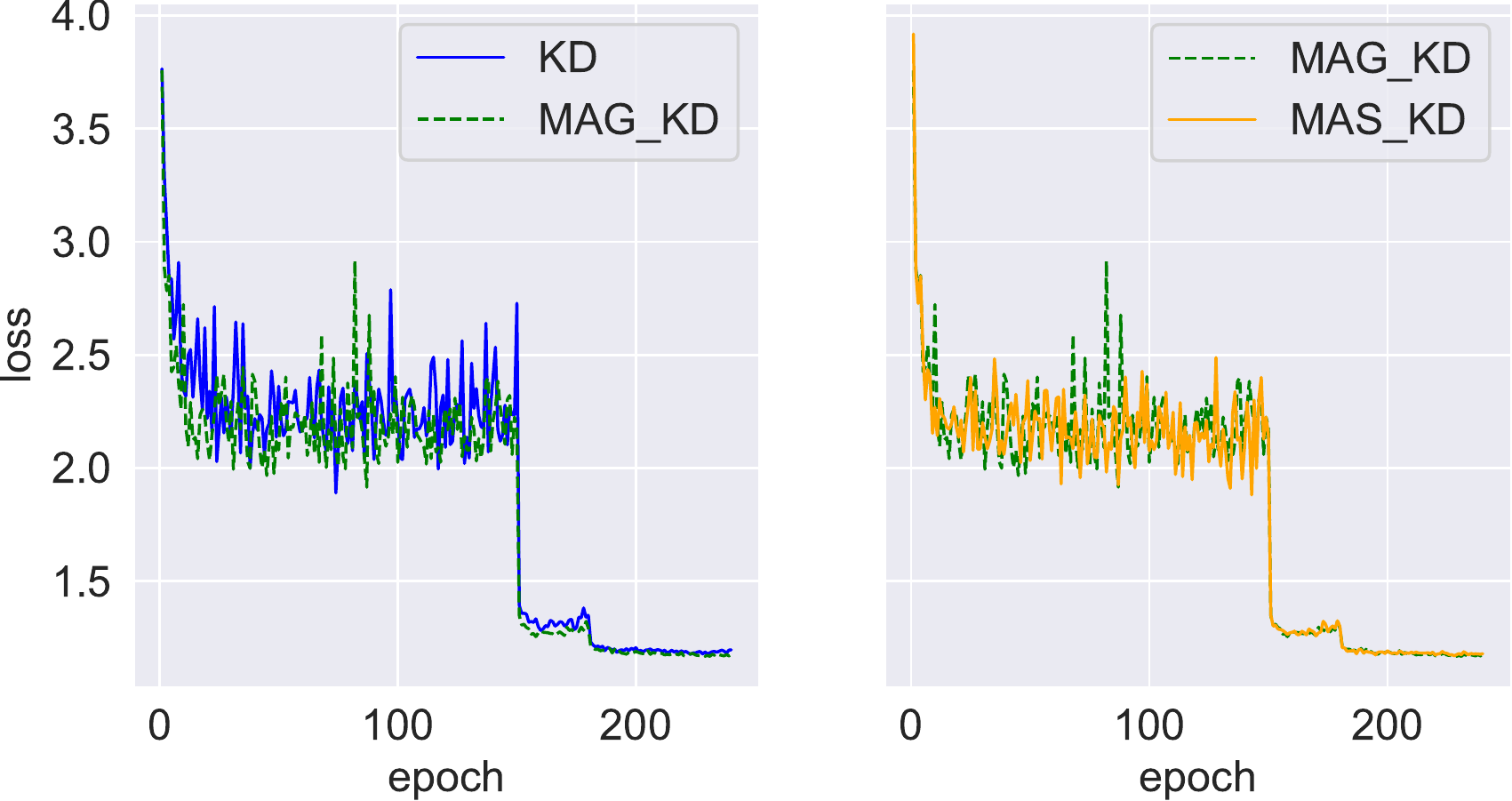}
	\caption{The figure shows the curves of the validation loss of HKD, MAG\_HKD and MAS\_HKD when vgg13 distills vgg8. To facilitate comparison, the figure is split into two subfigures, where the green dotted curve in the left figure is identical to the green dotted curve in the right one.}
	\label{fig:stable_loss_curve}
\end{figure}
For more stable teacher supervision during the early distillation period, a stable excitation scheme is designed for the distillation module, as shown in Figure \ref{fig:stable excitation mechanism}. The stable excitation distillation scheme integrates $F_{NK}^{(T_{SA})}$, $F_{AKB}^{(T_{SA})}$ and $F_{DKB}^{(T_{SA})}$, and distills the ensemble knowledge to $S$. The objective function is denoted as:
\begin{equation}
	L_{SE}=\sum_{k}L_H(F_{k}^{(T_{SA})}, F_{k}^{(S)}, \tau_k) + L_{EN} + L_{KD},
	\label{L_{SE}}
\end{equation}
where $k\in \left\{\mathrm{AK}, \mathrm{DK}\right\}$, $L_{EN}$ is denoted as:
\begin{equation}
	L_{EN}=L_{H}\left( F_{E}^{(T_{SA})},F_{NK}^{(S)},\tau_{NK} \right) ,
	\label{L_{EN}}
\end{equation}
where $\tau_{NK}$ is the native knowledge distillation temperature of $S$, $F_E^{(T_{SA})}$ is the result of the ensemble, which can be denoted as
\begin{equation}
	F_E^{(T_{SA})}=avg(F_{AKB}^{(T_{SA})},F_{NK}^{(T_{SA})},F_{DKB}^{(T_{SA})}).
	\label{avg}
\end{equation}
In order to clearly explain the two proposed distillation schemes, the module operation is given in Algorithm \ref{algorithm:distillation}. Contrastive analysis that vgg13 distills vgg8 on CIFAR-100 using HKD \cite{hinton2015KD} with different dillation schemes is conducted. It can be seen from Figure \ref{fig:stable_loss_curve} that the loss curves of both the HKD and MAG methods are jittery strongly during the early optimization period, and the loss curve of MAS is relatively stable, because the ensemble operation in MAS enables the teacher to provide robust supervision.
\begin{algorithm}
	\caption{Multi-granularity distillation}
	\label{algorithm:distillation}
	\KwIn{Frozen self-analyzed teacher $T_{SA}$, trainable student $S$ with AbstractedKE and DetailedKE.}
	\KwOut{Optimal student $S$.}
	\While{not converged}
	{
		Feed the image batch into $T_{SA}$ and $S$ respectively;\\
		Calculate the outputs of AbstractedKE, classifer and DetailedKE of $T_{SA}$: $F_{AK}^{(T_{SA})}$, $F_{NK}^{(T_{SA})}$ and $F_{DK}^{(T_{SA})}$;\\
		Calculate the outputs of AbstractedKE, classifer and DetailedKE of $S$: $F_{AK}^{(S)}$, $F_{NK}^{(S)}$ and $F_{DK}^{(S)}$;\\
		
		\uIf{Granularity-wise distillation}
		{
			Optimize $S$ via Equation \eqref{L_{GWD}}.
		}
		\ElseIf{Stable excitation distillation}
		{
			Calculate the outputs of AbstractedKB and DetailedKB of $T_{SA}$: $F_{AKB}^{T_{SA}}$ and $F_{DKB}^{T_{SA}}$;\\
			Calculate the ensemble of $F_{AKB}^{T_{SA}}$, $F_{NK}^{(T_{SA})}$ and $F_{DKB}^{T_{SA}}$ via Equation \eqref{avg};\\
			Optimize $S$ via Equation \eqref{L_{SE}} and \eqref{L_{EN}}.
		}
	}
\end{algorithm}

\section{Experiments}
In this section, we compare our method with state-of-the-art methods in knowledge distillation as well as online distillation. Some ablation studies are conducted to evaluate the effectiveness of designed scheme and sensitivity to hyper-parameters. A few further studies are also conducted to analyze the proposed multi-granularity knowledge and other performance.
\subsection{Datasets}
\label{Datasets}
Four classification datasets are used in our experiments, where CIFAR-100 and Market-1501 are used for model compression, STL-10 and TinyImageNet are used for representation transferring. \textbf{CIFAR-100} \cite{krizhevsky2009learning} is composed of $32\times32$ color images with 100 classes objects. \textbf{TinyImageNet} \cite{deng2009imagenet} consists of $64\times64$ images from 200 classes, each with 500 training images and 50 validation images. \textbf{STL-10} \cite{coates2011analysis} includes a training set of 5K labeled and 100K unlabeled $96\times96$ images from 10 classes and 8K test images. The Top-1 classification accuracy is usually adopted as evalution metric for the above datasets. \textbf{Market-1501} \cite{zheng2015scalable} is a dataset for person re-identification problem, which contains 12936 training images from 751 identities and 19732 test images from 750 identities. Rank-1 accuracy and mean average precision (mAP) metrics are reported.
\subsection{Implementation details}
\label{Implementation details}
For CIFAR-100, we conduct experiments using pretrained model provided by  Tian et al. \cite{tian2019crd} including VGG \citep{simonyan2014very}, ResNet \citep{he2016deep}, WideResNet \citep{zagoruyko2016wide}, MobileNet \citep{howard2017mobilenets} and ShuffleNet \citep{zhang2018shufflenet} as well as experimental settings. We use SGD with momentum for training granularity self-analyzing module and set the initial learning rate to 0.1, momentum to 0.9, minibatch to 64. All teacher models are trained for 60 epoches and the learning rate drops by 0.1 at 30, 45 epoch and all student models are trained for 240 epoches and the learning rate drops by 0.1 at 150, 180 and 210 epoch.
\\
For Market-1501, experiments are conducted on torchreid \cite{torchreid}. Adam is used for multi-granularity self-analyzing teacher training and we set the initial learning rate to 0.0015, momentum to 0.9, minibatch to 64. All teacher models are trained for 60 epoches and the learning rate drops by 0.1 at 30, 45 epoch and all student models are trained for 150 epoches and the learning rate drops by 0.1 every 60 epoches. \\
Parameters in multi-granularity mechanism are shown in Table \ref{tab:table_parameters}.
\begin{table}[]
	\caption{Parameter settings for multi-granularity mechanism. "Whole mechanism" denotes the parameters used in both multi-granularity self-analyzing module and distillation module.}
	\label{tab:table_parameters}
	\begin{tabular}{cccccccc}
		\toprule
		\multirow{2}{*}{Module}          & \multirow{2}{*}{Parameter} & \multicolumn{2}{c}{Value} \\ \cmidrule(l){3-4} 
		&                            & CIFAR-100  & Market-1501  \\ \midrule
		\multirow{2}{*}{Whole mechanism} & $dim_{AK}$                 & 64         & 256          \\
		& $dim_{DK}$                 & 256        & 1024         \\ \cmidrule(l){2-4} 
		\multirow{3}{*}{Self-analyzing}  & $\tau_{AKB}$               & 2.0        & 4.0          \\
		& $\tau_{NK}$                & 4.0        & 8.0          \\
		& $\tau_{DKB}$               & 8.0        & 12.0         \\ \cmidrule(l){2-4} 
		\multirow{3}{*}{Distillation}    & $\tau_{AK}$               & 2.0        & 4.0          \\
		& $\tau_{NK}$                & 4.0        & 8.0          \\
		& $\tau_{DK}$               & 8.0        & 12.0         \\ \bottomrule
	\end{tabular}
\end{table}
\subsection{Comparison with SOTAs}
\paragraph{Results on CIFAR-100} \label{Results on CIFAR100}
\begin{table*}[width=2.0\linewidth]
	\centering
	\caption{Top-1 accuracy (\%) of model compression on CIFAR-100. Bold indicates the best accuracy in each teacher-student pair, underline indicates the second best. Average over 3 runs.}
	\label{tab:results_CIFAR100}
	\begin{tabular}{cccccccc}
		\toprule
		Teacher      				     & WRN-40-2 & resnet56   & resnet110   & resnet32x4  & resnet32x4  & WRN-40-2   \\
		Student      					 & WRN-40-1 & resnet20   & resnet32    & ShuffleV2   & ShuffleV1  &  ShuffleV1  \\ \midrule
		Teacher     					 & 75.61    & 72.34      & 74.31       & 79.42  & 79.42  &    75.61\\
		Student      					 & 71.98    & 69.06      & 71.14       & 71.82  & 70.50  &    70.50\\ \midrule
		HKD \cite{hinton2015KD}     	 & 73.58    & 71.05      & 73.34       & 75.02  & 74.04  &    75.30\\
		FitNet \cite{Romero2015FitNetsHF} & 72.32    & 69.33      & 71.07       & 73.96  & 73.97  &    74.02\\
		AT \cite{komodakis2017paying}     & 72.82    & 70.39      & 72.60       & 73.13  & 72.20  &    74.20\\
		SP \cite{tung2019similarity}      & 73.00    & 70.28      & 72.74       & 74.96  & 74.31  &    75.24\\
		CC \cite{peng2019correlation}     & 71.93    & 69.76      & 71.65       & 72.13  & 71.39  &    71.74\\
		VID \cite{ahn2019variational}     & 73.23    & 70.53      & 72.67       & 73.62  & 73.95  &    74.05\\
		RKD \cite{park2019relational}     & 72.10    & 69.67      & 72.24       & 73.59  & 72.47  &    72.42\\
		PKT \cite{pkt_eccv}               & 73.48    & 70.62      & 72.57       & 75.07  & 74.19  &    74.39\\
		AB \cite{heo2019knowledge}        & 72.63    & 69.88      & 70.83       & 74.70  & 73.84  &    73.65\\
		FT \cite{kim2018paraphrasing}     & 71.41    & 69.99      & 72.44       & 73.49  & 72.23  &    72.87\\
		NST \cite{huang2017like}   		 & 72.37    & 69.75      & 71.79       & 74.68  & 74.12  &    74.89\\
		CRD \cite{tian2019crd}   		 & 74.24    & 71.38      & 73.55       & 75.95  & 75.20   &   76.05 \\
		AFD \cite{ji2021show}       	& 73.66    & 71.32      & 73.46       & 75.91  & 75.08  &    75.63\\ 
		WSL \cite{zhou2021wsl}          & 74.28    & \textbf{72.17}      & \textbf{74.06}       & 76.10  & \underline{75.46}   & 76.31 \\ \midrule
		MAG\_HKD    					 & 74.10    & 71.43      & 73.55       & 75.75  & 75.12  &    75.70\\
		MAS\_HKD   				        & 74.42    & 71.08      & 73.54       & 75.40  & 75.24  &    \underline{76.71}\\  
		MAG\_CRD                        & \underline{74.53}	& \underline{71.77}		& \underline{74.00}		&	\underline{76.37}	& 75.41	 & 76.14	\\
		MAS\_CRD                        & \textbf{74.80}	& 71.52		& \textbf{74.06}		& \textbf{77.14}		& \textbf{75.80} &  \textbf{77.13} \\
		\bottomrule
	\end{tabular}
\end{table*}
Model compression experiments are conducted on CIFAR-100, we rerun the code provided by Tian et al. \cite{tian2019crd} and report the results. For AFD \cite{ji2021show} which only conducted experiments on ResNet and WideResNet, we refer to the setting of FitNet \cite{Romero2015FitNetsHF} and VID \cite{ahn2019variational} for its compatibility with other network structures. Results are shown in Table \ref{tab:results_CIFAR100}, where MAG\_HKD means MAG embeded into HKD \cite{hinton2015KD}. It can be noticed that with the proposed MAG and MAS embeded, the methods overperforms SOTAs. For example, MAS\_CRD even helps ShuffleV1 exceed the accuracy of teacher WRN-40-2 by 1.52\%. It should be noticed that with the help of MAS, the HKD \cite{hinton2015KD}, whose performance was surpassed by CRD \cite{tian2019crd} by 0.64\% in average, outperforms CRD \cite {tian2019crd} by 0.28\% in vgg13-vgg8 pair, and by 0.66\% in WRN-40-2-ShuffleV1 pair. It can be seen from Table \ref{tab:results_CIFAR100} that in most cases, the proposed mechanism can process and extract high-level features of the image in terms of global and detailed aspects, obtain more discriminative features, and improve network performance. While for the resnet56-resnet20 network pair, the performance of the proposed method is worse than WSL \cite{zhou2021wsl}. It is maily due to the limited performance of the student resnet20 which is difficult to handle hard samples. The conventional distillation methods and the proposed mechanism do not consider this problem, so the performance is worse than WSL \cite{zhou2021wsl} which adaptively assigns proper weight for each sample.
\paragraph{Results on Market-1501} \label{Results on Market-1501}
\begin{table*}[width=2.0\linewidth]
	\centering
	\caption{Rank1 and mAP (\%) of models on Market-1501. Bold indicates the best performance. In the distillation settings, the models in the Setting 2 and 8 are used as the teacher networks of the Setting 11 and 12 respectively. And the results of Setting 9 and 10 are quoted from original papers.}
	\label{tab:Market1501}
	\begin{tabular}{cccccccc}
		\toprule
	Setting	& Method     & Backbone     & Rank1 & mAP   \\ \midrule
	1	&Vanilla     & ResNet50     & 88.84  & 71.59 \\
	2	&Vanilla     & DenseNet-121 & 90.17  & 74.02 \\
	3	&HA-CNN \cite{li2018harmonious}   	& Inception           & 90.90  & 75.60 \\ 
	4	&MLFN	\cite{chang2018multi}     & ResNeXt		 & 90.10 & 74.30  \\ 
	5	&PCB \cite{sun2018beyond}         & ResNet50     & 92.64  & 77.47 \\ 
	6	&Circle Loss \cite{sun2020circle} & DenseNet-121 & 91.00  & 76.54 \\
	7	&OSNet$\times0.75$ \cite{zhou2019omni} & OSNet   & 93.60 & 82.50 \\
	8	&OSNet$\times1.0$	\cite{zhou2019omni} & OSNet    & 94.10 & 82.90\\ 
	9	&CtF \cite{gong2020faster}		& Resnet50	& 93.70	& 84.00 \\ 
	10	& Auto-ReID \cite{gu2021auto}		& NAS	& 93.80	&	83.60 \\ \midrule
	11	&MAS\_RKD(T:2)     & ResNet50     & 91.09  & 79.43 \\ 
	12	&MAS\_RKD(T:8) & OSNet$\times0.75$ & \textbf{94.50} & \textbf{84.30} \\ 
		\bottomrule
	\end{tabular}
\end{table*}
Model compression experiments on Market-1501 are conducted as well. MAS combined with RKD \cite{park2019relational} (MAS\_RKD) is used in the setting, and results are shown in Table \ref{tab:Market1501}. It can be found that the proposed mechanism helps ResNet50 surpass its teacher DenseNet-121 in performance. The performance of the OSNet is also compared. Furthermore, the proposed mechanism increases the Rank1 of the OSNet$\times0.75$ by 0.9\% and mAP by 1.8\%, which means that the Rank1 of the OSNet$\times0.75$ exceeds the OSNet$\times1.0$ by 0.4\%, and the mAP exceeds the OSNet$\times1.0$ by 1.4\%.
\paragraph{Comparison with online distillation} \label{Online Distillation}
\begin{table*}[width=2.0\linewidth]
	\centering
	\caption{Test accuracy(\%) of online-distillation on CIFAR-100. We re-run DML using author-provided code, and reimplement KDCL based on the paper. $\uparrow$ denotes outperformance over SOTAs and $\downarrow$ denotes underperformance, bold denotes that both models in the pair have performance improvement.}
	\label{tab:results_onlinedistillation}
	\begin{tabular}{ccccccccccc}
		\toprule
		Model1                   & WRN-40-2 & resnet56 & resnet110 & resnet110 & resnet32x4       & ResNet50 & resnet32x4 & WRN-40-2 \\
		Model2                  & WRN-40-1 & resnet20 & resnet20  & resnet32  & resnet8x4     & vgg8     & ShuffleV1  & ShuffleV1  \\ \midrule
		\multirow{2}{*}{Vanilla}  & 75.61      & 72.34    & 74.31     & 74.31     & 79.42               & 79.34    & 79.42      & 75.61      \\
		                          & 71.98      & 69.06    & 69.06     & 71.14     & 72.50              & 70.36    & 70.50      & 70.50       \\ \midrule
		\multirow{2}{*}{DML \cite{zhang2018deep}}      & 76.36      & 72.88    & 73.98     & 74.22     & 79.43              & 79.07    & 79.96      & 76.45      \\
		                          & 71.54      & 69.18    & 69.92     & 71.76     & 73.19             & 70.57    & 71.72      & 72.04      \\ 
		\multirow{2}{*}{MAG\_DML}  & \textbf{76.78}$(\uparrow)$      & \textbf{73.51}$(\uparrow)$    & \textbf{74.44}$(\uparrow)$     & \textbf{74.25}$(\uparrow)$     & \textbf{79.88}$(\uparrow)$               & \textbf{79.22}$(\uparrow)$    & 79.86$(\downarrow)$      & 75.89$(\downarrow)$    \\
		                          & \textbf{72.06}$(\uparrow)$      & \textbf{69.77}$(\uparrow)$    & \textbf{70.11}$(\uparrow)$     & \textbf{72.08}$(\uparrow)$     & \textbf{73.20}$(\uparrow)$              & \textbf{71.27}$(\uparrow)$    & 72.29$(\uparrow)$      & 71.74$(\downarrow)$    \\ \midrule
		\multirow{2}{*}{KDCL \cite{guo2020online}}     & 78.07      & 74.56    & 76.10     & 75.96     & 80.44             & 80.38    & 80.48      & 77.64      \\
		                          & 73.01      & 70.70    & 70.19     & 72.42     & 74.39             & 72.44    & 74.42      & 74.59      \\ 
		\multirow{2}{*}{MAG\_KDCL} & \textbf{78.31}$(\uparrow)$      & \textbf{74.74}$(\uparrow)$    & \textbf{76.63}$(\uparrow)$     & \textbf{76.59}$(\uparrow)$     & 80.34$(\downarrow)$             & 79.35$(\downarrow)$    & \textbf{80.54}$(\uparrow)$      & \textbf{77.74}$(\uparrow)$      \\
                                  & \textbf{73.81}$(\uparrow)$      & \textbf{71.15}$(\uparrow)$    & \textbf{71.77}$(\uparrow)$     & \textbf{72.91}$(\uparrow)$     & 74.87$(\uparrow)$               & 72.67$(\uparrow)$    & \textbf{75.02}$(\uparrow)$      & \textbf{74.80}$(\uparrow)$       \\ \bottomrule
	\end{tabular}
\end{table*}
The proposed mechanism can be integrated into online distillation framework. The experiments are conducted on CIFAR-100 dataset, in which the settings are the same as the compression settings on CIFAR-100 and the accuracy of model pairs are reported. Table \ref{tab:results_onlinedistillation} shows the online distillation results. It can be noticed that MAG helps both DML and KDCL improve accuracy on six out of the eight model pairs. Compared with DML \cite{zhang2018deep}, MAG improves the accuracy of WRN-40-2-WRN-40-1 pair by 0.63\% and 0.52\% respectively. Compared with KDCL \cite{guo2020online}, MAG improves the accuracy of resnet110-resnet32 pair by 0.63\% and 0.83\% respectively. It can be seen from the table that for online distillation, multi-granularity knowledge also performes better than others in most cases, because multi-granularity knowledge is more abundant than conventional knowledge, and the information interaction between the two models is more sufficient.
\subsection{Ablation study}
Ablation study is conducted, including effectiveness of proposed multi-granularity self-analyzing module and distillation schemes, and sensitivity to hyper-parameters.
\paragraph{Effectiveness of multi-granularity self-analyzing}  
\label{Effectiveness of self-analyzing}
\begin{figure*}
	\centering
	\subfigure[output distribution at logits]{
		\includegraphics[width=0.25\linewidth]{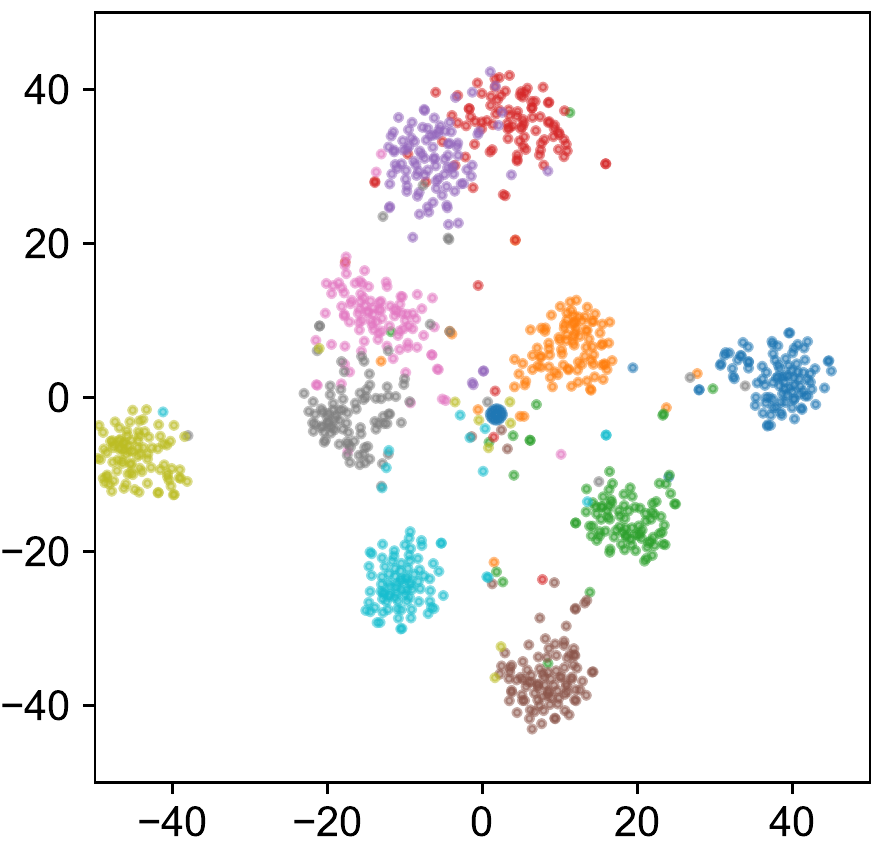}
		\label{fig:tsne-logits}
	}%
	\centering
	\subfigure[output distribution of AbstractedKE]{
		\includegraphics[width=0.25\linewidth]{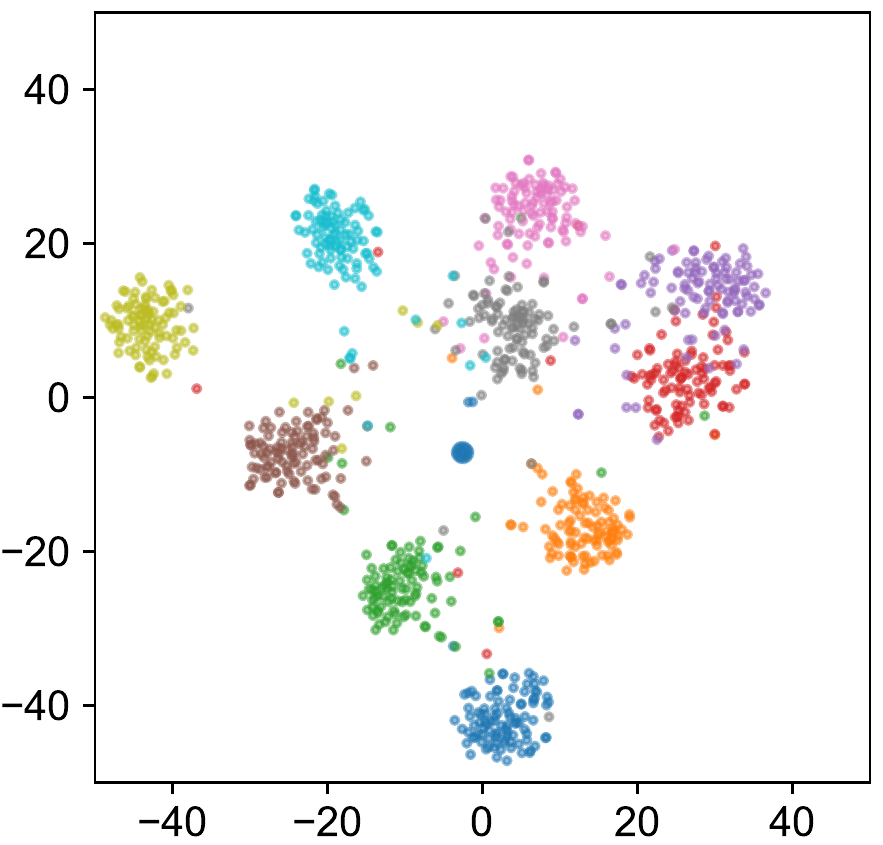}
		\label{fig:tsne-low}
	}%
	\centering
	\subfigure[output distribution of DetailedKE]{
		\includegraphics[width=0.25\linewidth]{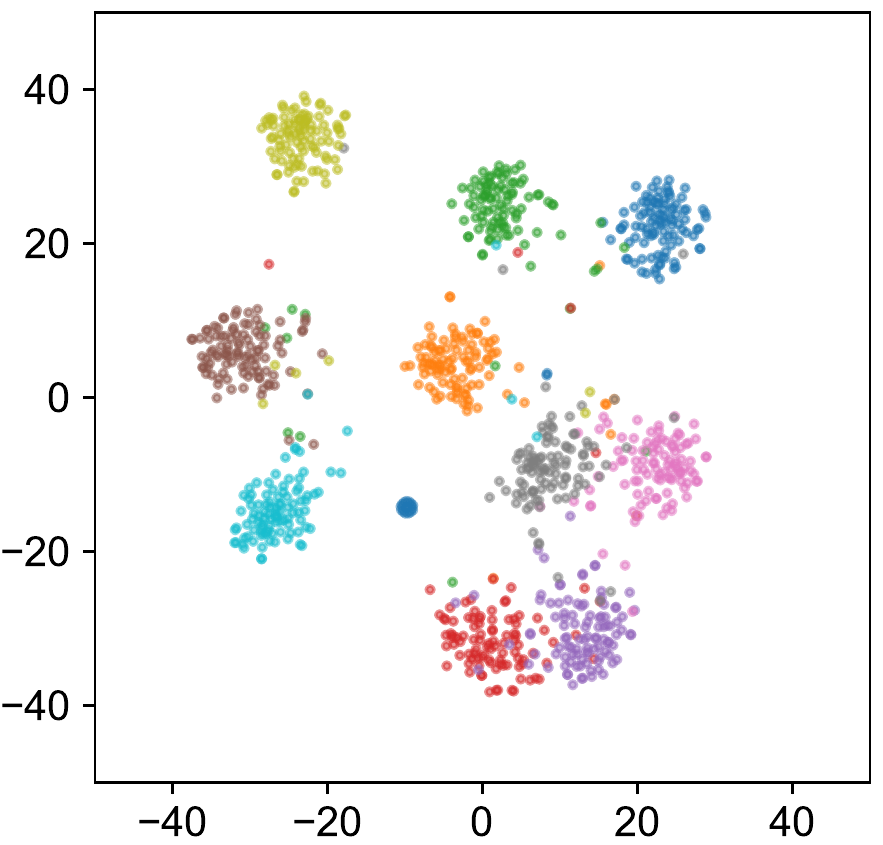}
		\label{fig:tsne-high}
	}%
	\caption{The output distribution of resnet56 after multi-granularity self-analyzing. The first subfigure on the left illustrates the logit distribution of the model and the two subfigures on the right illustrate the output distribution of corresponding encoder. Each point in the figure represents a sample, and different colors represent different categories.}
	\label{fig:tsne_analysis}
\end{figure*}
We use T-SNE to visualize the output distribution of resnet56 after multi-granularity self-analyzing, as shown in Figure \ref{fig:tsne_analysis}. Ten categories are randomly selected from CIFAR-100 validation set for drawing. For fairness and convenience, we limit the scale of the three subfigure to a fixed size of $100\times100$. The figure shows that the AbstractedKE and the DetailedKE retain the classification representation of the network's classifier, which is a prerequisite for the proposed mechanism to play a role. It can be also found that the distribution of logit overlaps slightly, and the samples represented by orange are not fully clustered, which easily lead to misclassification. The problems are alleviated by the two encoders from which some samples that are linearly inseparable in the native dimension may become linearly separable because of the dimensional transformation.
\paragraph{Effectiveness comparison of MAG and MAS}~{}\\
(1) Knowledge distillation between networks with similar structure. The experiment conducted to demonstrate the effectiveness of the proposed mechanism and schemes, the results are reported in Table \ref{tab:ablation_CIFAR100_similarity_structure}. It can be seen that the proposed mechanism improves the performance of SOTAs on almost all teacher-student pairs. MAG and MAS improve the accuracy of RKD \cite{park2019relational} by 2.2\% and 2.36\% on resnet32x4-resnet8x4 pair respectively, and improve the accuracy of SP \cite{tung2019similarity} by 0.5\% and 1.49\% on resnet110-resnet20 pair respectively. The improvement effect of the proposed mechanism on AFD is not very satisfied, because the constraints imposed on students by AFD and the proposed may make the network fail to optimize to the optimal solution.\\
(2) Knowledge distillation between networks with different structure. The experiment conducted to demonstrate the effectiveness of the proposed mechanism and schemes as well, the results are reported in Table \ref{tab:ablation_CIFAR100_different_structure}. It can be seen that the proposed mechanism improves the performance of SOTAs on almost all teacher-student pairs. MAG and MAS improve the accuracy of AT \cite{komodakis2017paying} by 2.39\% and 9.07\% on ResNet50-MobileV2 pair respectively, and improve the accuracy of CRD \cite{tian2019crd} by 0.42\% and 1.19\% on resnet32x4-ShuffleV2 pair respectively.
\begin{figure}[]
	\centering
	\subfigure[Accuracy under different dimensions of encoders]{
		\includegraphics[width=0.8\linewidth]{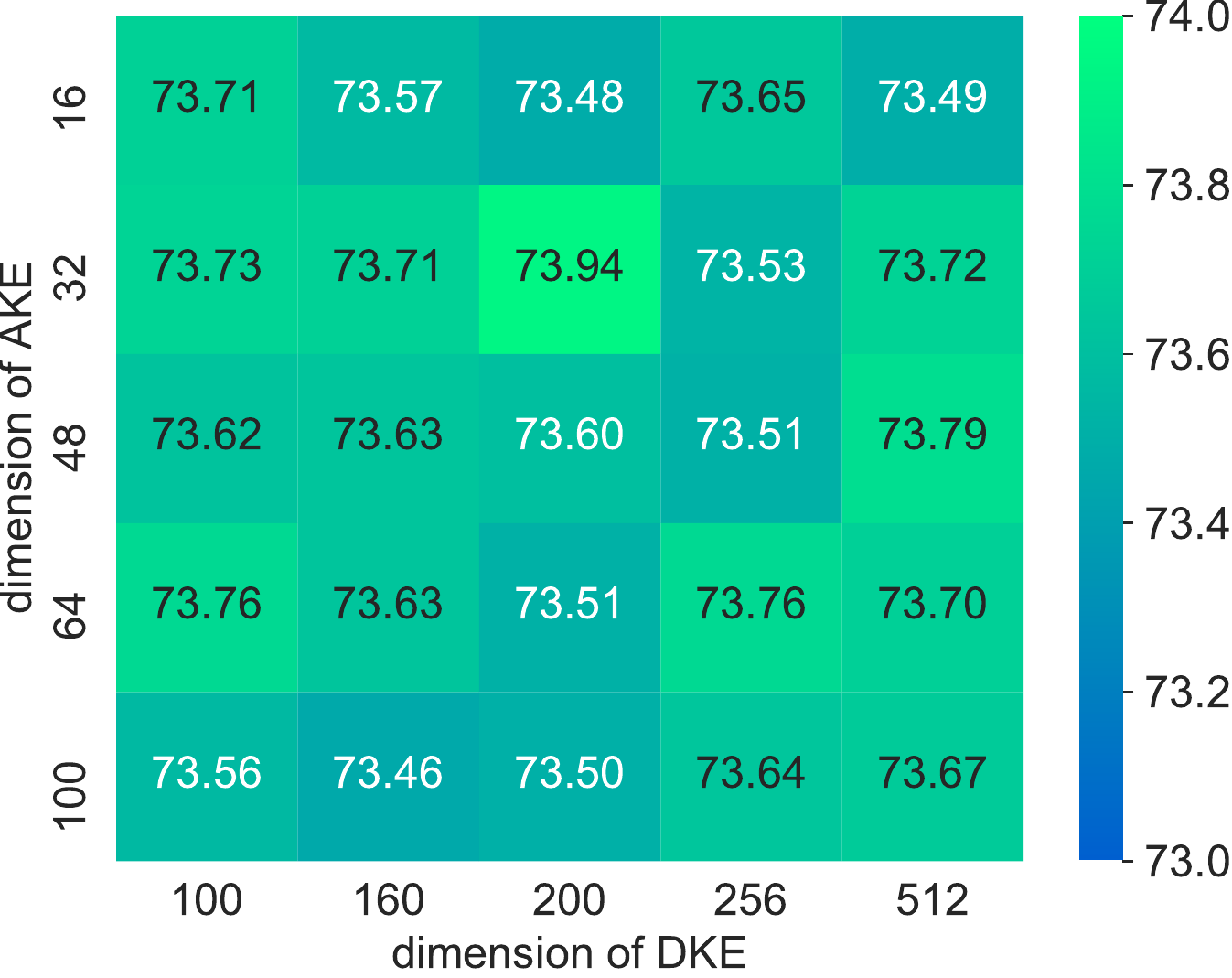}
		\label{fig:ablation_dim}
	}%
	\\
	\subfigure[Accuracy under different branch distillation temperatures]{
		\includegraphics[width=0.8\linewidth]{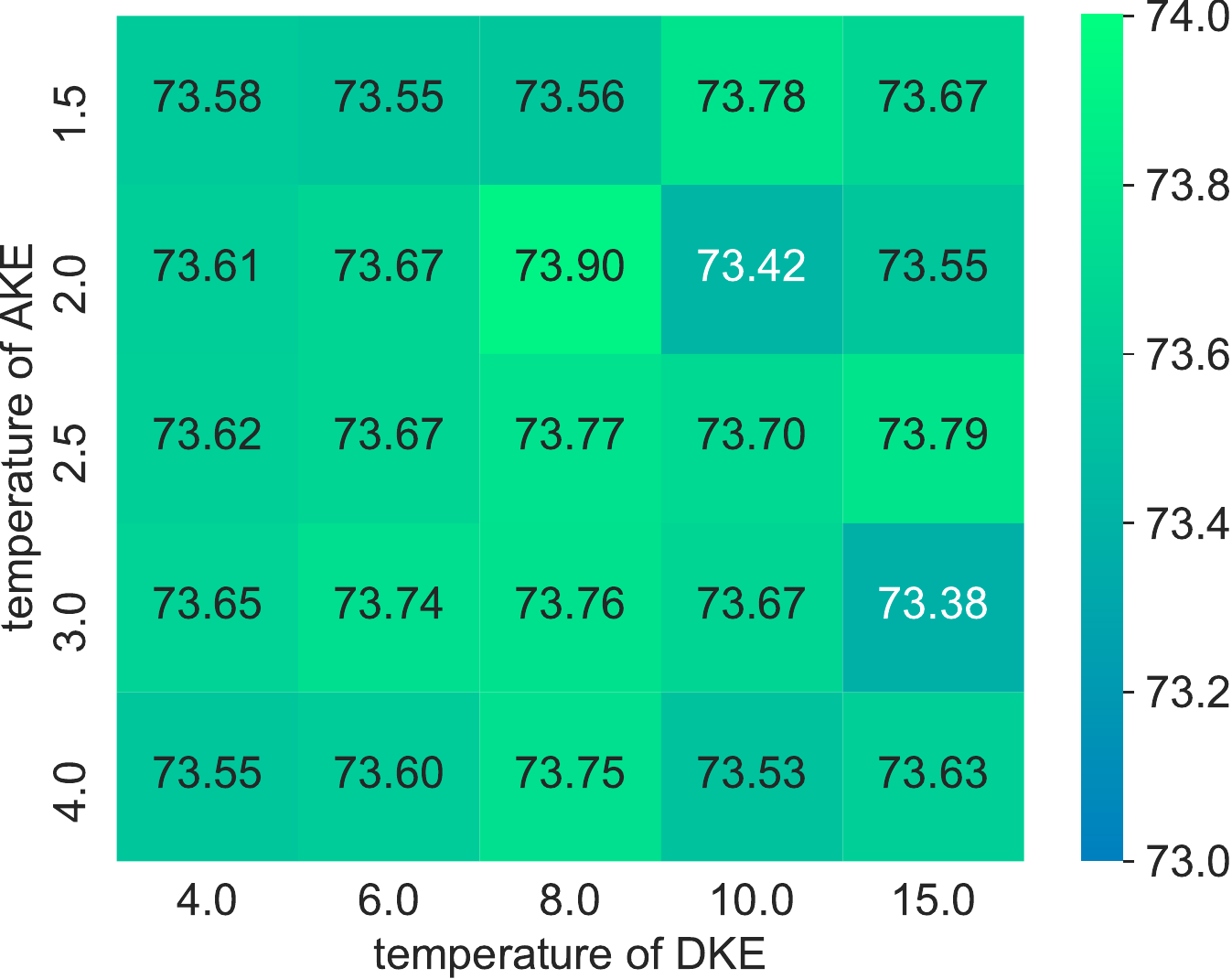}
		\label{fig:ablation_temperature}
	}%
	\caption{Ablation study about dimensions and temperature of two encoders on CIFAR-100 with MAG\_HKD. The teacher is vgg13 and the student is vgg8. Average over 2 runs.}
	\label{fig:abaltion}
\end{figure}
\begin{table*}[width=2.0\linewidth]
	\centering
	\caption{Ablation study of similiar teacher-student structure model compression on CIFAR-100. $\uparrow$ denotes outperformance over SOTAs. Bold indicates the best accuracy in each pair with different methods. Average over 3 runs.}
	\label{tab:ablation_CIFAR100_similarity_structure}
	\begin{tabular}{lccccccc}
		\toprule
		Teacher      & WRN-40-2    & WRN-40-2 &resnet56 & resnet110 & resnet110 & resnet32x4  & vgg13\\
		Student      & WRN-16-2 & WRN-40-1 & resnet20     & resnet20  & resnet32  & resnet8x4 & vgg8 \\ \midrule
		Teacher      & 75.61    & 75.61 & 72.34 & 74.31 & 74.31 & 79.42  & 74.64\\
		Student      & 73.26 & 71.98 & 69.06     & 69.06  & 71.14  & 72.50 & 70.36 \\ \midrule
		HKD \cite{hinton2015KD}       & 73.59             & 73.58             & 71.05             & 70.90              & 73.34             & 73.27             & 73.18 \\
		+MAG  & 75.09$(\uparrow)$ & 74.10$(\uparrow)$ & \textbf{71.43}$(\uparrow)$ & \textbf{71.53}$(\uparrow)$  & \textbf{73.55}$(\uparrow)$ & 73.82$(\uparrow)$ & 73.63$(\uparrow)$ \\
		+MAS & \textbf{75.35}$(\uparrow)$ & \textbf{74.42}$(\uparrow)$ & 71.08$(\uparrow)$ & 71.03$(\uparrow)$ & 73.54$(\uparrow)$      & \textbf{74.20}$(\uparrow)$      &  \textbf{74.18}$(\uparrow)$     \\     \midrule
		FitNet \cite{Romero2015FitNetsHF}   & 73.82             & 72.32             & 69.33             & 68.96              & 71.07             & 73.62              & 71.14 \\
	+MAG  & \textbf{75.30}$(\uparrow)$ & 73.99$(\uparrow)$ & 70.29$(\uparrow)$ & 70.31$(\uparrow)$  & 72.73$(\uparrow)$ & 74.88$(\uparrow)$  & 73.06$(\uparrow)$ \\
	+MAS & 75.17$(\uparrow)$      & \textbf{74.43}$(\uparrow)$      & \textbf{71.08}$(\uparrow)$      &  \textbf{70.69}$(\uparrow)$      & \textbf{73.18}$(\uparrow)$      & \textbf{75.76}$(\uparrow)$      & \textbf{73.59}$(\uparrow)$      \\     \midrule
		AT \cite{komodakis2017paying}       & 74.39             & 72.82             & 70.39             & 70.36              & 72.60             & 73.53              & 71.41 \\
	+MAG      & 75.28$(\uparrow)$ & 73.81$(\uparrow)$ & 70.99$(\uparrow)$ & 70.57$(\uparrow)$  & \textbf{73.56}$(\uparrow)$ & 74.56$(\uparrow)$  & 72.11$(\uparrow)$ \\
	+MAS     & \textbf{75.98}$(\uparrow)$      & \textbf{74.90}$(\uparrow)$      & \textbf{71.78}$(\uparrow)$      &  \textbf{71.34}$(\uparrow)$      & 73.29$(\uparrow)$      & \textbf{74.92}$(\uparrow)$      & \textbf{73.38}$(\uparrow)$      \\     \midrule
		SP \cite{tung2019similarity}       & 74.01             & 73.00             & 70.28             & 70.29              & 72.74             & 73.28              & 72.94 \\
	+MAG      & 74.30$(\uparrow)$ & 73.71$(\uparrow)$ & \textbf{71.13}$(\uparrow)$ & 70.79$(\uparrow)$ & 73.44$(\uparrow)$ & 73.58$(\uparrow)$ & 73.20$(\uparrow)$ \\
	+MAS & \textbf{75.37}$(\uparrow)$      & \textbf{73.79}$(\uparrow)$      & 70.97$(\uparrow)$      &  \textbf{71.78}$(\uparrow)$      &  \textbf{73.66}$(\uparrow)$     & \textbf{74.26}$(\uparrow)$      & \textbf{73.64}$(\uparrow)$      \\     \midrule
		VID \cite{ahn2019variational}      & 74.19             & 73.23             & 70.53             & 70.68              & 72.67             & 73.24              & 71.41 \\
	+MAG     & 74.84$(\uparrow)$ & 73.35$(\uparrow)$& 71.14$(\uparrow)$ & 70.69$(\uparrow)$  & 73.00$(\uparrow)$ & 74.73$(\uparrow)$  & 72.92$(\uparrow)$ \\
	+MAS & \textbf{75.63}$(\uparrow)$      &\textbf{74.49}$(\uparrow)$     & \textbf{71.28}$(\uparrow)$      &  \textbf{71.61}$(\uparrow)$      & \textbf{73.32}$(\uparrow)$      & \textbf{74.86}$(\uparrow)$     &  \textbf{73.56}$(\uparrow)$     \\     \midrule
		RKD \cite{park2019relational}      & 73.37             & 72.10             & 69.67             & 69.44              & 72.24             & 72.03              & 71.35 \\
    +MAG& \textbf{75.73}$(\uparrow)$      & 73.59$(\uparrow)$     &  71.51$(\uparrow)$     & \textbf{71.11}$(\uparrow)$      & \textbf{73.71}$(\uparrow)$    &  74.23$(\uparrow)$     & \textbf{73.44}$(\uparrow)$      \\     
    +MAS & 75.31$(\uparrow)$      & \textbf{74.30}$(\uparrow)$      &  \textbf{71.91}$(\uparrow)$     & 71.06$(\uparrow)$       &  73.17$(\uparrow)$     & \textbf{74.39}$(\uparrow)$      & 73.06$(\uparrow)$      \\     \midrule
	CRD \cite{tian2019crd}      & 75.52 & 74.24 & 71.38 & 71.34 & 73.55 & 75.32  & 73.90  \\
	+MAG& 75.84$(\uparrow)$ & 74.53$(\uparrow)$ & \textbf{71.77}$(\uparrow)$ & \textbf{71.91}$(\uparrow)$ & 74.00$(\uparrow)$ & \textbf{75.89}$(\uparrow)$ & \textbf{74.29}$(\uparrow)$ \\ 
	+MAS & \textbf{75.87}$(\uparrow)$      & \textbf{74.80}$(\uparrow)$      &  71.52$(\uparrow)$     & 71.52$(\uparrow)$       &  \textbf{74.06}$(\uparrow)$     & 75.41$(\uparrow)$      &  74.06$(\uparrow)$   \\ \midrule
	AFD \cite{ji2021show}      & 75.41  & 73.66 & 71.32 & 71.20 & 73.46 & 74.72 & 73.57 \\
	+MAG & 75.53$(\uparrow)$ & \textbf{74.53}$(\uparrow)$ & \textbf{71.62}$(\uparrow)$ & \textbf{71.40}$(\uparrow)$ & 73.57$(\uparrow)$ & 74.75$(\uparrow)$ & \textbf{73.89}$(\uparrow)$ \\
	+MAS  & \textbf{75.55}$(\uparrow)$ & 74.12$(\uparrow)$ & 71.49$(\uparrow)$ & 71.22$(\uparrow)$ & \textbf{74.00}$(\uparrow)$  & \textbf{75.03}$(\uparrow)$ & 73.62$(\uparrow)$ \\
	\bottomrule
	\end{tabular}
\end{table*}
\begin{table*}[width=2.0\linewidth]
	\centering
	\caption{Ablation study of different teacher-student structure model compression on CIFAR-100. $\uparrow$ denotes outperformance over SOTAs. Bold indicates the best accuracy in each pair with different methods. Average over 2 runs.}
	\label{tab:ablation_CIFAR100_different_structure}
	\begin{tabular}{lcccccccc}
		\toprule
		Teacher      & vgg13    & ResNet50 & ResNet50 & resnet32x4 & resnet32x4 & WRN-40-2  \\
		Student      & MobileV2 & MobileV2 & vgg8     & ShuffleV1  & ShuffleV2  & ShuffleV1 \\ \midrule
		Teacher      & 74.64    & 79.34    & 79.34    & 79.42      & 79.42      & 75.61     \\
		Student      & 64.60    & 64.60    & 70.36    & 70.50      & 71.82      & 70.50      \\ \midrule
		HKD \cite{hinton2015KD}           & 67.69    & 68.18    & 73.94    & 74.04      & 75.02      & 75.30      \\
		+MAG      & 68.36$(\uparrow)$    & \textbf{69.00}$(\uparrow)$    & 74.02$(\uparrow)$    & 75.12$(\uparrow)$      & \textbf{75.75}$(\uparrow)$      & 75.70$(\uparrow)$      \\
		+MAS     & \textbf{68.68}$(\uparrow)$    & 68.58$(\uparrow)$    & \textbf{74.11}$(\uparrow)$    & \textbf{75.24}$(\uparrow)$      & 75.40$(\uparrow)$     &  \textbf{76.71}$(\uparrow)$          \\ \midrule
		FitNet \cite{Romero2015FitNetsHF}       & 63.91    & 63.23    & 69.93    & 73.97      & 73.96      & 74.02     \\
		+MAG  & 66.49$(\uparrow)$    & 65.74$(\uparrow)$    & 71.35$(\uparrow)$    & 75.73$(\uparrow)$      & 76.32$(\uparrow)$      & 76.04$(\uparrow)$     \\
		+MAS & \textbf{67.51}$(\uparrow)$    & \textbf{68.40}$(\uparrow)$    & \textbf{73.10}$(\uparrow)$    & \textbf{76.81}$(\uparrow)$ & \textbf{76.60}$(\uparrow)$  &\textbf{76.56}$(\uparrow)$           \\ \midrule
		AT \cite{komodakis2017paying}           & 59.77    & 58.82    & 72.02    & 72.20       & 73.13      & 74.20      \\
		+MAG     & 62.13$(\uparrow)$    & 61.21$(\uparrow)$    & 72.96$(\uparrow)$    & 74.20$(\uparrow)$       & 75.02$(\uparrow)$      & 75.73$(\uparrow)$     \\
		+MAS     & \textbf{66.76}$(\uparrow)$    & \textbf{67.89}$(\uparrow)$    & \textbf{74.37}$(\uparrow)$    & \textbf{75.87}$(\uparrow)$  & \textbf{76.17}$(\uparrow)$  & \textbf{76.79}$(\uparrow)$          \\ \midrule
		SP \cite{tung2019similarity}           & 66.64    & 67.77    & 73.58    & 74.31      & 74.96      & 75.24     \\
		+MAG      & 67.63$(\uparrow)$    & 68.43$(\uparrow)$    & \textbf{74.26}$(\uparrow)$    & 75.66$(\uparrow)$      & 76.54$(\uparrow)$      & 76.09$(\uparrow)$     \\
		+MAS     & \textbf{69.50}$(\uparrow)$         &  \textbf{68.93}$(\uparrow)$        & 73.83$(\uparrow)$         &  \textbf{75.73}$(\uparrow)$          & \textbf{76.47}$(\uparrow)$           &  \textbf{76.26}$(\uparrow)$         \\ \midrule
		VID \cite{ahn2019variational}          & 65.80    & 67.57    & 70.60    & 73.95      & 73.62      & 74.05     \\
		+MAG     & 67.74$(\uparrow)$    & 67.64$(\uparrow)$    & 72.98$(\uparrow)$    & 75.06$(\uparrow)$      & 76.01$(\uparrow)$      & 76.26$(\uparrow)$     \\
		+MAS    & \textbf{69.47}$(\uparrow)$         & \textbf{69.01}$(\uparrow)$         & \textbf{73.55}$(\uparrow)$         & \textbf{75.61}$(\uparrow)$           &  \textbf{76.81}$(\uparrow)$          &  \textbf{76.35}$(\uparrow)$         \\ \midrule
		RKD \cite{park2019relational}          & 64.62    & 64.43    & 71.68    & 72.47      & 73.59      & 72.42     \\
		+MAG     & \textbf{68.77}$(\uparrow)$    & 68.04$(\uparrow)$    & 73.98$(\uparrow)$    & 75.05$(\uparrow)$      & 75.05$(\uparrow)$      & 76.17$(\uparrow)$     \\
		+MAS    & 68.29$(\uparrow)$         &   \textbf{68.94}$(\uparrow)$       & \textbf{74.14}$(\uparrow)$         &  \textbf{75.69}$(\uparrow)$          &   \textbf{76.02}$(\uparrow)$         &  \textbf{76.33}$(\uparrow)$         \\ \midrule
		CRD \cite{tian2019crd}          & 69.69    & 69.11    & 74.31    & 75.20      & 75.95      & 76.05     \\
		+MAG     & \textbf{69.80}$(\uparrow)$    & 69.63$(\uparrow)$    & 74.56$(\uparrow)$    & 75.41$(\uparrow)$      & 76.37$(\uparrow)$      & 76.14$(\uparrow)$     \\
		+MAS    &  69.78$(\uparrow)$        & \textbf{70.03}$(\uparrow)$         &  \textbf{74.84}$(\uparrow)$        & \textbf{75.80}$(\uparrow)$          &  \textbf{77.14}$(\uparrow)$          &  \textbf{77.13}$(\uparrow)$ \\        \midrule
		AFD \cite{ji2021show}     & 68.97  & 69.13 & 73.37 & 75.08 & 75.91 & 75.63 \\
		+MAG & \textbf{69.45}$(\uparrow)$ & \textbf{69.49}$(\uparrow)$ & 73.68$(\uparrow)$ & 75.29$(\uparrow)$ & 76.39$(\uparrow)$ & 75.73$(\uparrow)$ \\
		+MAS  & 69.32$(\uparrow)$ & 69.40$(\uparrow)$ & \textbf{73.95}$(\uparrow)$ & \textbf{75.43}$(\uparrow)$ & \textbf{76.57}$(\uparrow)$  & \textbf{76.21}$(\uparrow)$ \\
		\bottomrule
	\end{tabular}
\end{table*}
\paragraph{Sensitivity to encoder dimensions}
\label{Sensitivity to encoder dimensions}
We set the dimension of DetailedKE to 100, 160, 200, 256 and 512, and the dimension of AbstractedKE to 16, 32, 48, 64 and 100 on CIFAR-100 respectively. Figure \ref{fig:ablation_dim} shows the results of sensitivity to encoder dimensions. The accuracy ranges from 73.46 to 73.94, which means that the mechanism is not sensitive to the dimensions of the encoders. 
\paragraph{Sensitivity to encoder temperatures}
\label{Sensitivity to encoder temperatures}
We set the temperature of DetailedKE to 4.0, 6.0, 8.0, 10.0, 15.0, and the temperature of AbstractedKE to 1.5, 2.0, 2.5, 3.0, 4.0 on CIFAR-100 respectively. Figure \ref{fig:ablation_temperature} shows the results of sensitivity to encoder distillatin temperature. The accuracy ranges from 73.38 to 73.90, which means that the proposed mechanism is not sensitive to the distillation temperature of the encodes.
\subsection{Further study}
\begin{figure}[]
	\centering
	\includegraphics[width=0.8\linewidth]{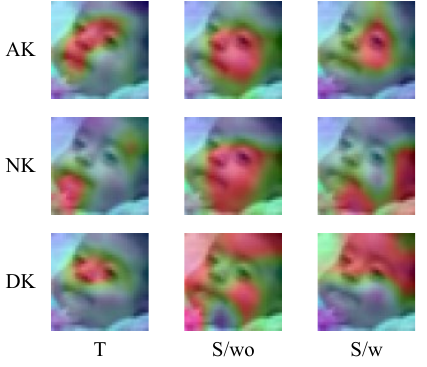}
	\caption{CAMs of teacher resnet32x4, student ShuffleV1 trained without and with the proposed distillation. The left, middle, and the right column show CAMs of multi-granularity knowledge of the three networks respectively. AK, NK and DK denote abstracted, native and detailed knowledge respectively.}
	\label{fig:cam}
\end{figure}

\paragraph{Visualization of multi-granularity knowledge}
Class activation map (CAM) \cite{zhou2016learning} is utilized to visualize multi-granularity knowledge and its transferability. CAM of teacher resnet32x4 with multi-granularity via self-analyzing is shown in the left column of Figure \ref{fig:cam}, and CAMs of student ShuffleV1 trained without and with distillation are shown in the right two columns of Figure \ref{fig:cam}. \\
It can be noticed that for the teacher network, abstracted knowledge focuses on the almost entire face. detailed knowledge tends to pay attention to the details of the face, such as the eyes and nose, and native knowledge concerns about the forehead and mouth, which are less discriminative than eyes and nose. Because of the different effectiveness of multi-granularity knowledge focusing on the discriminative areas, the proposed multi-granularity knowledge will work.
\\
It can be also found from the Figure \ref{fig:cam} that three CAMs of the student trained with the proposed distillation (S/w) are closer to those of the teacher than those of the student trained without the distillation (S/wo), which means the proposed multi-granularity knowledge has been transferred. Spectially, the detailed knowledge of S/wo in Figure \ref{fig:cam} almost focuses on the entire picture, including background areas, while detailed knowledge of S/w reduces the focus on the background and mouth, and increases the focus on the eyes and nose. We will quantitatively analyze the transferability of multi-granularity knowledge in the knowledge similarity of similarity analysis experiment.

\begin{figure}[]
	\centering
	\includegraphics[width=0.8\linewidth]{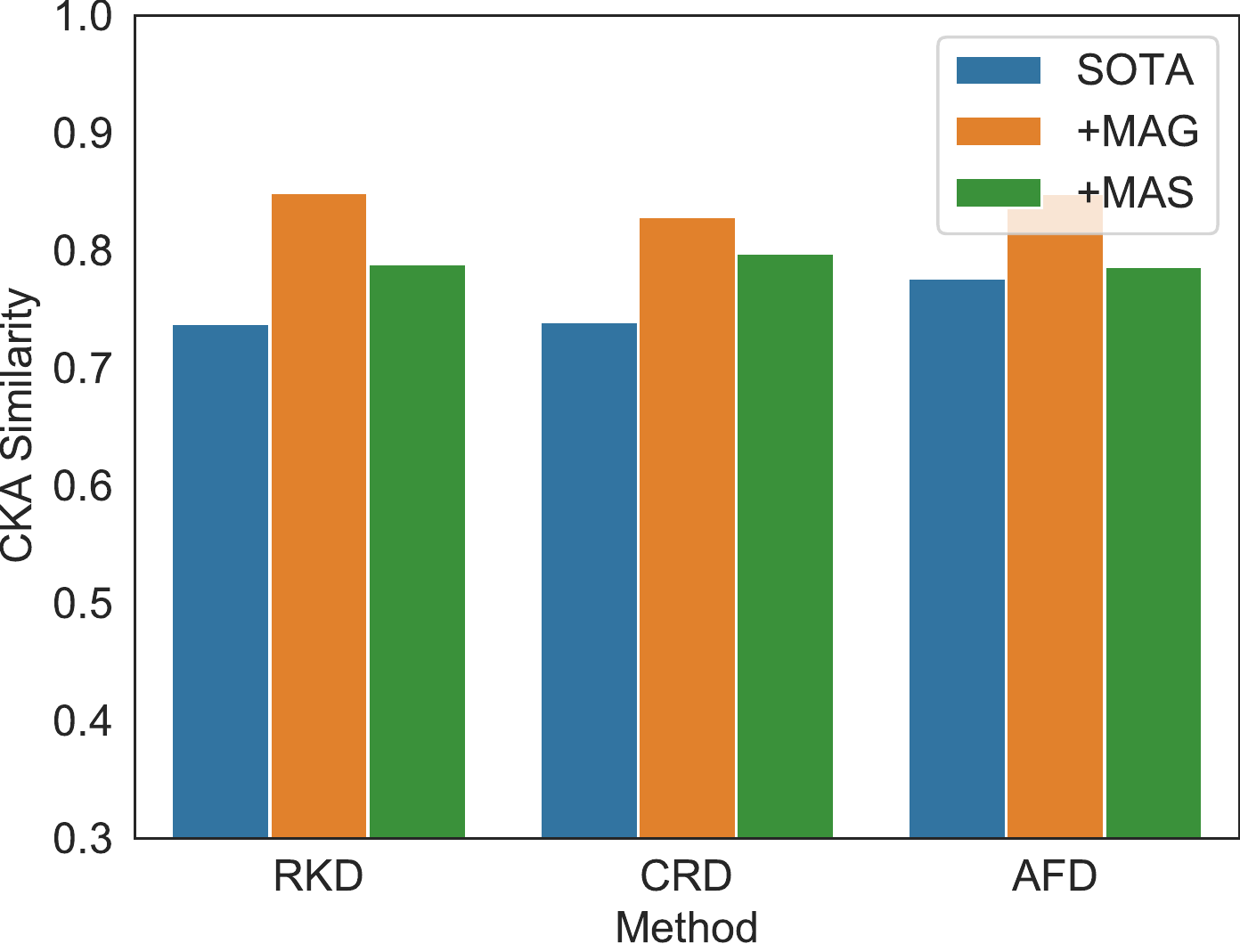}
	\caption{CKA-similarity between resnet32x4 and ShuffleV2 using three distillation methods on CIFAR-100. RBF CKA-similarity results are reported.}
	\label{fig:cka_analysis}
\end{figure}

\begin{figure}[]
	\centering
	\includegraphics[width=\linewidth]{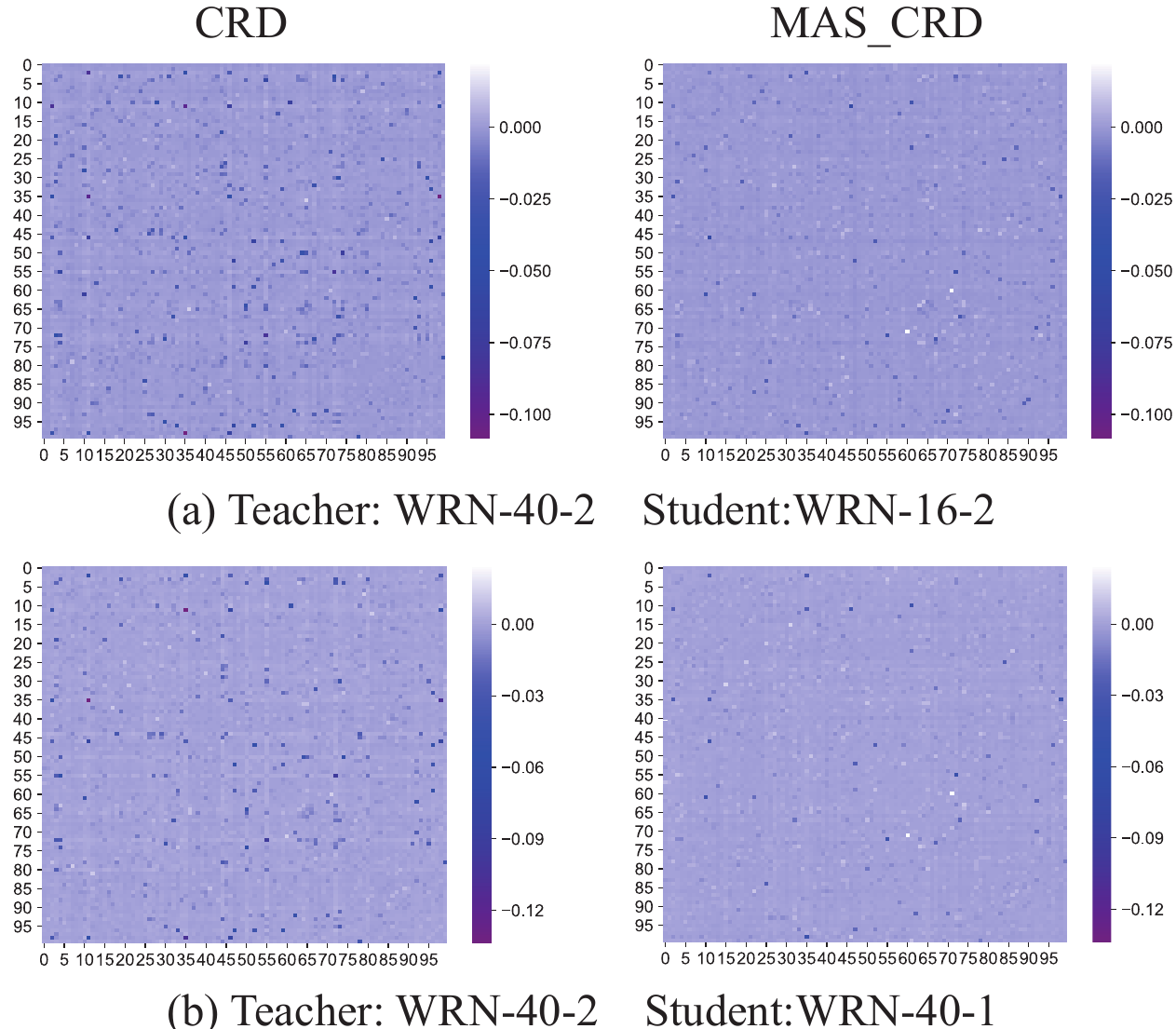}
	\caption{The difference between correlation matrices of the teacher's and the student's logits on CIFAR-100. The left column shows the result of CRD method \cite{tian2019crd}, and the right column shows the result of our MAS\_CRD method. The closer the difference is to 0, the better the information between the classes captured by the student.}
	\label{fig:correlation}
\end{figure}

\begin{table}[]
	\centering
	\caption{Quantative analysis of multi-granularity knowledge transferring performance on CIFAR-100 validation set. $\uparrow$ denotes the larger the value, the better, while $\downarrow$ denotes that the smaller the value, the better. AK denotes abstracted knowledge, NK denotes native knowledge and DK denotes detailed knowledge.}
	\label{tab:transferablity}
	\begin{tabular}{@{}ccccccc@{}}
		\toprule
		& \multicolumn{3}{c}{w/o distillaion} & \multicolumn{3}{c}{with distillation} \\ \cmidrule(l){2-7} 
		& AK         & NK        & DK         & AK          & NK         & DK         \\ \midrule
		SSIM($\uparrow$)           & 0.003     & 0.205    & -0.001    & 0.656      & 0.484     & 0.747     \\
		cosine($\uparrow$) & -0.001    & 0.722    & 0.012    & 0.946      & 0.871     & 0.970     \\
		Pearson($\uparrow$)        & 0.007     & 0.783    & -0.002    & 0.868      & 0.816     & 0.921     \\
		L2($\downarrow$)             & 0.671     & 0.260    & 0.070     & 0.073      & 0.245     & 0.006     \\ \bottomrule
	\end{tabular}
\end{table}
\paragraph{Similarity analysis}~{}\\
(1) Knowledge similarity. In order to quantify the transferability of multi-granularity knowledge, four metrics relative to the teacher are used to evaluate the student's outputs of AbstractedKE, classifier and DetailedKE, namely structural similarity (SSIM) \cite{wang2004image}, cosine similarity \cite{cosine}, Pearson correlation coefficient \cite{benesty2009pearson} and L2 distance. For SSIM \cite{wang2004image}, the outputs of AbstractedKE, classifier and DetailedKE are vectorized. The results are reported in Table \ref{tab:transferablity}. It can be found that the knowledge of the three granularities of the student trained with distillation is better than that trained without distillation on the four evaluation metrics, which proves the transferability of the multi-granularity knowledge we proposed.\\
(2) $T/S$ network similarity. Generally, the higher the similarity between the student and the teacher, the better the performance of the student can be ensured. We calculate the similarity between resnet32x4 and ShuffleV2 with CKA-similarity \cite{kornblith2019similarity} on CIFAR-100 validation set. The proposed MAG and MAS improve the CKA-similarity compared to the three SOTA methods, illustrated as Figure \ref{fig:cka_analysis}. Interestingly, we find that the accuracy of MAS is higher than that of MAG in most cases, which can be seen from Table \ref{tab:ablation_CIFAR100_different_structure}, while the CKA-similarity of MAS is lower than MAG. The reason can be as follows. MAG directly uses the teacher's native knowledge as the supervision of the student native knowledge, while MAS takes the integrated knowledge of the teacher as the supervision of the student native knowledge. The integrated knowledge is biased compared with the teacher's native knowledge itself, so the similarity between the student and the teacher distilled with MAS is lower than that with MAG. \\
(3) Visualization of correlation matrice difference. The correlation between the probabilities of the network for each category is called dark knowledge \cite{hinton2015KD}. Following CRD \cite{tian2019crd}, correlation matrice difference between the teacher and the student at logits on CIFAR-100 are visualized, as shown in Figure \ref{fig:correlation}, from which we can see that the proposed mechanism helps CRD \cite{tian2019crd} capture dark knowledge that is closer to that in the teacher network.\\
From the results of the above analysis experiments, it can be seen that multi-granularity knowledge can be transferred, and multi-granularity knowledge distillation can improve the network similarity between the student and the teacher. The visual effect of the correlation matrice difference obtained by using the proposed multi-granularity knowledge distillation is better.

\paragraph{Transferability of representations}
\label{Transferability of representations}
Fine-tuning with pretrained network on target dataset produces a boost to generalization \cite{yosinski2014transferable}. We transfer the CIFAR-100 pretrained resnet20 distilled by resnet56 to STL-10 and TinyImageNet, by freezing the backbone and fine-tuning the classfier. The transfer performance results can be seen in Table \ref{tab:result_transfer}. The proposed MAG improves the transferring performance on nine out of ten settings. Compared with SP \cite{tung2019similarity} transferring to STL-10, MAG improves the transfer performance of resnet20 by 1.9\%. Compared with SP \cite{tung2019similarity} transferring to TinyImageNet, MAG improves the transfer performance of resnet20 by 2.7\%.
\begin{table*}[width=2.0\linewidth]
	\centering
	\caption{Transfer learning from CIFAR-100 to STL-10 and TinyImageNet. The transfer performance of MAG and SOTAs are compared. Bold denotes the better accuracy. C100 is the abbreviation of CIFAR-100, S10 is the abbreviation of STL-10 and TI is the abbreviation of TinyImageNet.}
	\label{tab:result_transfer}
	\begin{tabular}{@{}c|ccccc|ccccccc@{}}
		\toprule
		&  HKD   & FitNet & SP   & CRD    &  AFD  & MAG\_HKD         & MAG\_FitNet     & MAG\_SP            & MAG\_CRD & MAG\_AFD       \\ \midrule
		C100$\rightarrow$S10       &  69.1 & 68.1   & 68.0 & 70.2 & 69.7  & \textbf{69.8} & \textbf{68.5} & \textbf{69.9}  & \textbf{70.7}  & \textbf{70.0}\\
		C100$\rightarrow$TI & 30.4    &  29.5   & 30.2  & 34.2 & 32.6  & \textbf{32.6} & \textbf{32.0} & \textbf{32.9}  & \textbf{34.3} & 32.6\\ \bottomrule
	\end{tabular}
\end{table*}

\begin{figure}[]
	\centering
	\subfigure[Accuracy change under different noisy input]{
		\includegraphics[width=0.7\linewidth]{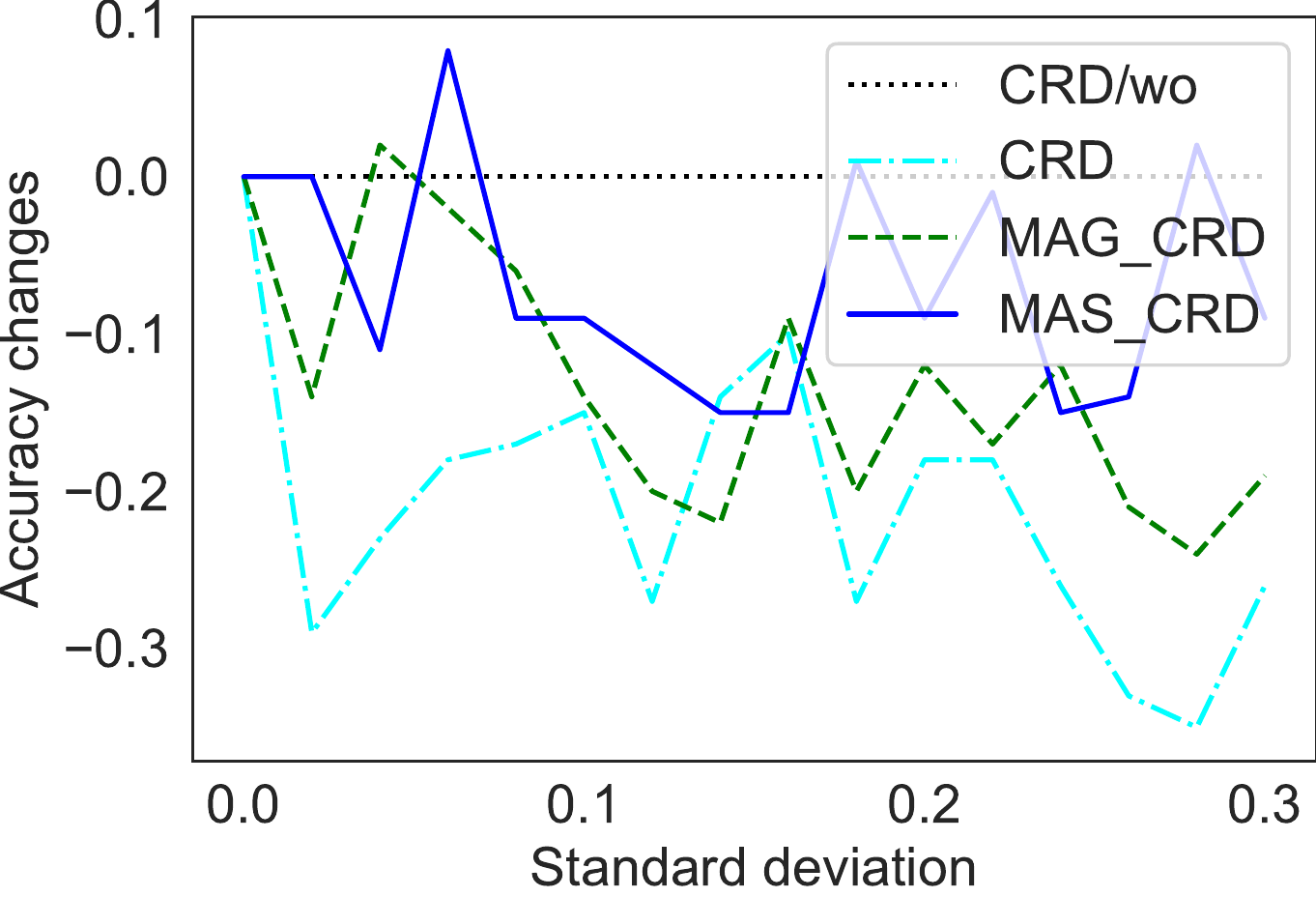}
		\label{fig:noise-crd}
	}%
	\\
	\subfigure[Variance of the accuracy change]{
		\includegraphics[width=0.7\linewidth]{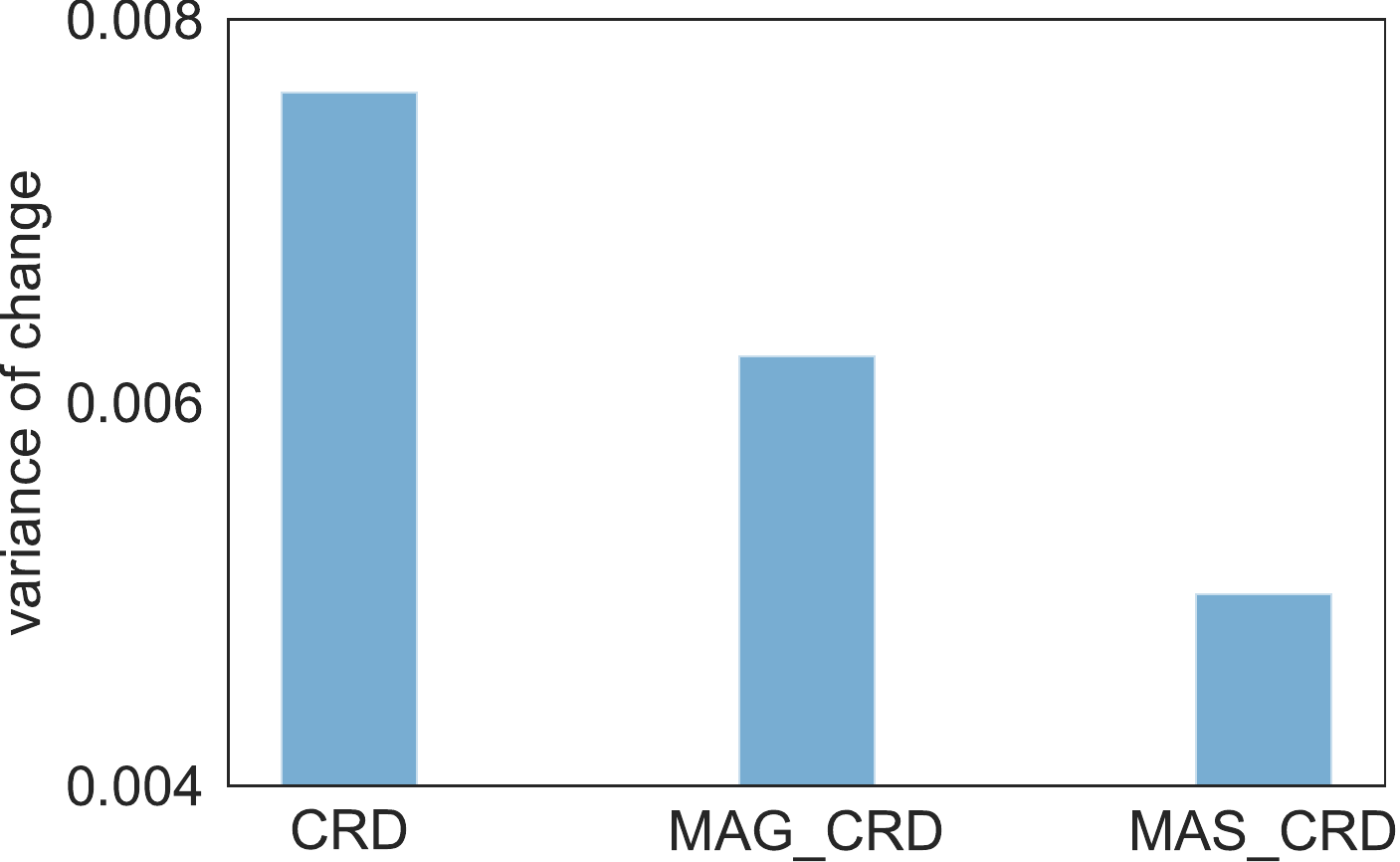}
		\label{fig:noise-crd-var}
	}%
	\caption{We evaluate the robustness to noise of different distillation methods. The teacher is ResNet50 and the student is MobileNetV2. In \subref{fig:noise-crd}, positive change denotes accuracy improvement and negative change denotes accuracy decrease. In \subref{fig:noise-crd-var}, the smaller the variance, the better.}
	\label{fig:noise}
\end{figure}

\paragraph{Robustness to noisy inputs}\label{Robustness to noise}
The robustness to noisy images of the proposed MAG and MAS and SOTA methods are compared. Gaussian noise is added to the CIFAR-100 validation images and the accuracy of the student MobileNetV2 is evaluated, where gaussian noise obeys $N(0, \sigma^2)$ normal distribution and $\sigma$ ranges from 0 to 0.3 by step 0.02. Accuracy changes relative to the settings with no added noise and the variance of the change are reported, as shown in Figure \ref{fig:noise}, where the black horizontal dotted line represents CRD without noisy input (CRD/wo). Figure \ref{fig:noise} shows that the accuracy of CRD \cite{tian2019crd} decrease when taking noisy images as the input, and the proposed MAS and MAG help the distillation be robust to the noise. It is interesting to observe that with the help of MAS, sometimes the accuracy of the distilled student with noisy inputs is even higher than that with noise-free inputs. Besides, the proposed MAG and MAS decrease the variance of the accuracy change in the settings, as illustrated in Figure \ref{fig:noise-crd-var}. It is confirmed that the proposed mechanism achieves better anti-noise ability.

\section{Conclusion}
A multi-granularity distillation mechanism is proposed for transferring multi-granularity knowledge which is easier for student networks to understand. To enable the teacher to construct multi-granularity knowledge, a granularity self-analyzing module of teacher is designed. In order to optimize the student under robust supervision, a stable excitation scheme is designed, which integrates the constructed knowledge. The proposed mechanism can be embedded into SOTA distillation frameworks to further improve the accuracy. It also improves the student transfer ability and robustness to noisy inputs.


\bibliographystyle{cas-model2-names}

\bibliography{reference}

\end{document}